\documentclass{article}
\usepackage[final, nonatbib]{neurips_2023}

\usepackage[utf8]{inputenc} 
\usepackage[T1]{fontenc}    
\usepackage{hyperref}       
\usepackage{url}            
\usepackage{booktabs}       
\usepackage{amsfonts}       
\usepackage{nicefrac}       
\usepackage{microtype}      
\usepackage[table,xcdraw]{xcolor}         
\usepackage{multirow}
\usepackage{algorithm}  
\usepackage{algpseudocode}  
\usepackage{amsmath}  
\usepackage{tabularx}
\usepackage{graphicx}
\usepackage{tablefootnote}
\usepackage{bbm}
\usepackage{float}
\usepackage{overpic}
\usepackage{soul,color,xcolor}
\usepackage{todonotes}
\usepackage{mathtools}
\usepackage{wrapfig}
\usepackage{subfigure}

\usepackage[numbers]{natbib}

\title{Exploring Time Granularity on Temporal Graphs\\for Dynamic Link Prediction in Real-world Networks }

\author{%
  Xiangjian Jiang\thanks{Equal contribution.} \\
  Department of Computer Science and Technology \\
  University of Cambridge\\
  \texttt{xj265@cam.ac.uk} \\
  \And
  Yanyi Pu$^*$ \\
  Department of Computer Science\\
  University of Sheffield \\
  \texttt{ypu17@sheffield.ac.uk} \\
}

\usepackage{comment}

\begin{document}

\maketitle

\begin{abstract}
\looseness-1
Dynamic Graph Neural Networks (DGNNs) have emerged as the predominant approach for processing dynamic graph-structured data. However, the influence of temporal information on model performance and robustness remains insufficiently explored, particularly regarding how models address prediction tasks with different time granularities. In this paper, we explore the impact of time granularity when training DGNNs on dynamic graphs through extensive experiments. We examine graphs derived from various domains and compare three different DGNNs to the baseline model across four varied time granularities. We mainly consider the interplay between time granularities, model architectures, and negative sampling strategies to obtain general conclusions. Our results reveal that a sophisticated memory mechanism and proper time granularity are crucial for a DGNN to deliver competitive and robust performance in the dynamic link prediction task. We also discuss drawbacks in considered models and datasets and propose promising directions for future research on the time granularity of temporal graphs. Our benchmark suite and codebase are available at {\color[HTML]{0070C0} \url{https://github.com/SilenceX12138/Time-Granularity-on-Temporal-Graphs}}.
\end{abstract}

\section{Introduction}
Evolving connections and relationships pervade real-world scenarios, encompassing recommendation systems~\cite{wu2022graph, gao2023survey}, social networks~\cite{yang2022s}, transportation systems~\cite{zhao2019t, zhang2019spatial, zhou2020variational}, epidemic transmission~\cite{wang2022causalgnn, adiga2022ai}, and more~\cite{longa_graph_2023}. Temporal graphs (also known as dynamic or evolutionary networks) can effectively model these dynamics, with nodes, edges, edge types, and associated attributes continuously changing over time due to various events. Analysing temporal evolution and its patterns can facilitate reliable predictions and informed decisions~\cite{longa_graph_2023, chen2021learning}. In contrast to traditional graph-based models that assume fixed graph structures, dynamic graph neural network methods (DGNNs) have emerged in recent years to enable more efficient representation learning on dynamic networks~\cite{skarding2021foundations}.

Time granularity significantly influences the level of detail at which temporal information is captured, processed, and represented in a model~\cite{chen2021learning, mao2019spatio, mubashar2022efficient}. It critically affects model performance, robustness, computational efficiency, and transferability. However, its impact on temporal graph analysis remains under-explored. We underscore the importance of understanding the time granularity of temporal graphs in the following aspects: \textbf{(i) Model Performance}: Coarser time granularities may sacrifice critical temporal information, whereas finer granularities could introduce noise into the training process~\cite{qian2021exploring}. Identifying the optimal choice of time granularity for a specific task can enhance model performance. \textbf{(ii) Robustness}: Assessing models at various time granularities enables evaluation of their robustness to information loss, which is crucial for determining how models generalise across time scales~\cite{qian2021exploring, chen2021learning} and their sensitivity to the provided temporal information. \textbf{(iii) Computational Efficiency}: Fine-grained models can be computationally intensive and slow to train due to numerous training instances; coarser granularities can reduce computational demands while retaining valuable insights~\cite{kazemi2020representation}. Balancing granularity and efficiency can expedite the temporal graph analysis process. \textbf{(iv) Transferability}: Knowledge acquired at different time granularities can be transferred between domains and tasks~\cite{qian2021exploring, su2020transferable, su2021transferable}. Hence, understanding the impact of time granularity could assist researchers in selecting and adapting models for specific problems.

\looseness-1
In this study, we focus on a fundamental task in dynamic graph analysis: dynamic link prediction. This task seeks to predict future connections and interactions based on prior and current information. While state-of-the-art (SOTA) methods~\cite{rossi2020temporal, kumar2019predicting, wang2021inductive} can achieve near-perfect performance on this task, previous studies~\cite{poursafaei_towards_2022} have shown that incorporating more challenging negative sampling techniques significantly reduces the performance of existing SOTA models. We extend this research by considering the impact of time granularity. Our primary objective is to provide a comprehensive understanding of how these models process temporal information and address the link prediction task in the absence of sufficiently fine-grained information. Our main contributions are as follows:

\begin{itemize}
    \item We introduce a novel data-splitting approach that jointly considers durations and the coarsest common time granularities across different dynamic graphs. This framework ensures a fair comparison among various graphs without information leakage issues.
    \item We empirically investigate DGNNs' performance and robustness under four predetermined time granularities. We conduct a series of controlled experiments, taking into account model architectures, dataset domains, and negative sampling strategies. 
    \item We perform cross-granularity evaluations on trained models across different time granularities to gain a deeper understanding of the models' mechanisms for processing temporal information at various granularity levels. 
    \item We provide an insightful discussion on the identified problems and innate weaknesses for the model design and datasets. 
\end{itemize}

\section{Related Work}

\looseness-1
\paragraph{Dynamic Graph Representation Learning}
\vspace{-2mm}
Over the past decade, learning effective representations that capture both structural properties and temporal information in dynamic graphs has been extensively studied, driven by the growing interest in temporal graphs. Several survey papers~\cite{kazemi_representation_2020, skarding_foundations_2021, longa_graph_2023} document the advancements in this research field, offering various taxonomies for classifying different types of dynamic graphs based on network characteristics and domains. Numerous Dynamic Graph Neural Networks (DGNNs) have been proposed to address long-standing challenges in representation learning: (i) incorporating diverse types of events, and (ii) integrating temporal information into node, edge, or graph embeddings. These DGNNs can be broadly categorised into four groups: Recurrent Neural Network (RNN)-based methods~\cite{ma2018streaming, kumar2019predicting}, memory-based methods~\cite{pareja2019evolvegcn, wang2022inductive}, attention-based methods~\cite{fathy_temporalgat_2020, sankar2019dynamic, sankar_dysat_2020}, and convolution-based methods~\cite{cong2023dyformer, cui2021dygcn}. In this study, we mainly focus on the Temporal Graph Networks (TGNs) because they generalise Message Passing Neural Networks (MPNNs) to temporal graphs with effective memory mechanism~\cite{rossi2020temporal}.

\vspace{-3mm}
\looseness-1
\paragraph{Time Granularity for Graphs}
Time granularity, which refers to the temporal resolution or time intervals at which dynamic graphs are observed or analyzed, significantly impacts the effectiveness of dynamic graph analysis by determining the level of temporal detail retained. Chapter 3 of the Handbook of Temporal Reasoning in AI~\cite{EuzenatMontanari2005} delivers a comprehensive introduction to time granularity and its applications in various domains. The authors present a mathematical formalization of the concept and provide a thoughtful discussion on the changes in semantic meanings of objects or concepts due to varying time granularities. In video representation learning, one pioneering work addresses the problem of choosing temporal granularity with advanced architectures. Specifically,   Qian et al.~\cite{qian2021exploring} demonstrate that proper granularity is task-dependent and coarse-grained features can be effective for many tasks where fine-grained data are considered to be necessary.

The study of time granularity in dynamic graphs has attracted increasing attention in recent years~\cite{huang2023temporal, poursafaei_towards_2022}. The pioneering work of Holme and Saramäki~\cite{holme_temporal_2012} in 2012 initiated the exploration of graph time granularity, examining the trade-offs between coarse and fine temporal resolutions. Concurrently, Casteigts et al.~\cite{hutchison_expressivity_2013} investigated the impact of time granularity on graph structures and expressivity, evaluating scenarios with varying waiting times for information transition. Additionally, Skarding et al.~\cite{skarding_foundations_2021} introduced a taxonomy for representing various temporal graphs across different time granularities based on link duration. In recent years, several studies~\cite{10.1145/3184558.3191526, kazemi2019time2vec, xu2019selfattention, fathy_temporalgat_2020} have presented novel time-aware embedding methods, primarily aimed at transforming static embeddings into continuous embeddings using temporal information with the consideration on different time intervals. This study aims to provide some insights into the impact of time granularity on learning representations from temporal graphs by focusing on the dynamic link prediction task.

\section{Notations and Formalism} \label{notation}


A dynamic graph $\mathbf{G_D}$, also known as a temporal graph, can be perceived as a sequence of operations guided by an ordered series of events acting upon the initial graph state $\mathbf{G_0}$ (a Continuous-time Dynamic Graph $\mathbf{G_{CT}}$), or a stream of snapshots over predetermined time slices (a Discrete-time Dynamic Graph $\mathbf{G_{DT}}$). A continuous-time temporal graph can be formally defined as $\mathbf{G_{CT}} = (G_0, O)$, where $G_0$ represents the initial state and $O = \{(u_i, v_i, x_i, t_i, \Delta{i}), i = 1, 2, ...\}$ indicates a sequence of events occurring at arbitrary time point. In this notation, $(u_i, v_i)$ represents the pair of nodes involved in the $i$-th event, $x_i$ denotes any feature associated with the event, including node and edge features, event types (communication, interaction, or other topological changes), and associated operations such as node/edge addition or deletion, $t_i$ marks the event's initiation timestamp, and $\Delta{i}$ represents the event's duration. In contrast, a discrete-time temporal graph can be expressed as an ordered sequence of $N$ graph snapshots $\mathbf{G_{DT}} = (G_0, G_1, ..., G_N)$. Each $G_i$ in the sequence may capture a snapshot of the graph at a specific time point $t_i$ or model the graph's state during the time interval $[t - \Delta t, t]$ for a configurable $\Delta t$. In the latter case, $\mathbf{G_i} = \big(\oplus V_i, \oplus E_i, \oplus X_i, (t_i - t_{i-1}) \big)$ offers an aggregated view of the dynamic graph $\mathbf{G_{TD}}$ over the specified time interval, with $\oplus$ denoting any possible method of combining events occurring within that time frame. The time interval may be fixed to represent different time granularities or adjusted over time to create overlapping snapshots, thereby facilitating more sophisticated analyses of the dynamic network.

In line with the data processing procedure in Poursafaei et al.'s work~\cite{poursafaei_towards_2022}, we eliminate all node and edge features from the graph data and focus exclusively on edge additions while assuming their permanence. This approach streamlines the events to be $O = \{(u_i, v_i, t_i), i = 1, 2, ...\}$. Furthermore, by permitting multiple identical edges between any pair of nodes, we can express any $G_i \in G_{TD}$ as:


\begin{equation}
\begin{split}
    \oplus E_i &= \oplus E_{i-1} \uplus \{(u, v) \mid \exists t \in (t_{i-1}, t_i] \text{ s.t. } (u, v, t) \in O \} \\
    \oplus V_i &= \cup \{ u, v \mid (u, v) \in \oplus E_i\},
\end{split}
\end{equation}

where $\uplus$ is the union operation on multisets, signifying the multiplicities of edges in $\oplus E_i$; $\cup$ is the standard union operation for the set of vertices. This equation represents the aggregated view of the graph from time $t_{i-1}$ to $t_i$, making it equivalent to an instantaneous snapshot of the graph at time $t_i$.

\section{Methods}

\subsection{Datasets}
In this study, we examine seven datasets of varying sizes from diverse domains, including social, interaction, and proximity. These datasets are widely employed for training and evaluating dynamic graph neural networks (DGNNs). Each dataset comprises directed edges listed by Unix timestamp, without any associated node or edge features. Additionally, the number of nodes in these datasets remains constant over time, simplifying the research scenario to focus exclusively on dynamic link predictions.Table~\ref{tab:dataset_sum} in Appendix~\ref{appendix:real_data} provides detailed statistics, as well as the semantic meanings of nodes and edges for each dataset.

\begin{table}[!thbp]
\centering
\caption{Datasets split by "Day" with the split rate of 2/3-1/6-1/6 with respect to the number of days for training, validation, and testing.}
\label{tab:data_split}
\resizebox{\textwidth}{!}{
\begin{tabular}{lrrrrrrrr}
\toprule
\multicolumn{1}{c}{\textbf{}} & \multicolumn{2}{c}{\textbf{Train}}                               & \multicolumn{2}{c}{\textbf{Validation}}                          & \multicolumn{2}{c}{\textbf{Test}}                                & \multicolumn{2}{c}{\textbf{Total}}                               \\
\toprule
\textbf{Dataset}              & \multicolumn{1}{c}{\# Days} & \multicolumn{1}{c}{\# Edges} & \multicolumn{1}{c}{\# Days} & \multicolumn{1}{c}{\# Edges} & \multicolumn{1}{c}{\# Days} & \multicolumn{1}{c}{\# Edges} & \multicolumn{1}{c}{\# Days} & \multicolumn{1}{c}{\# Edges} \\
\midrule
Wikipedia            & 20                             & 99,701                          & 5                              & 26,697                          & 5                              & 26,359                          & 30                             & 152,767                         \\
Reddit               & 20                             & 432,543                         & 5                              & 110,004                         & 5                              & 126,518                         & 30                             & 669,075                         \\
MOOC                 & 20                             & 216,364                         & 5                              & 65,815                          & 5                              & 63,421                          & 30                             & 345,610                         \\
LastFM               & 1,216                          & 916,312                         & 304                            & 340,736                         & 305                            & 26,566                          & 1,825                          & 1,284,223                       \\
Enron                & 730                            & 6,224                           & 182                            & 6,357                           & 183                            & 10,051                          & 1,095                          & 22,997                          \\
Social Evo.          & 160                            & 268,758                         & 40                             & 136,849                         & 40                             & 160,325                         & 240                            & 566,012                         \\
UCI                 & 130                            & 55,202                          & 32                             & 2,402                           & 34                             & 1,307                           & 196                            & 58,977 \\       \bottomrule             
\end{tabular}}
\vspace{-5mm}
\end{table}

To process the datasets, we first follow prior work~\cite{poursafaei_towards_2022} and aggregate their edges from the finest time granularity (second) to coarser granularities (minute, hour, day, month, year), and then adopt a 0.7-0.15-0.15 chronological split for training, validation, and test sets. The comprehensive breakdown is available in Appendix~\ref{data_distribution}. Our analysis reveals that "Day" represents the coarsest time granularity in this study, as it can meaningfully split all datasets without altering the semantic meanings of the original data. In contrast, the other coarser granularities might lead to insufficient samples of validation/test splits, e.g., the "Month" granularity of the Wikipedia dataset. 

Therefore, we choose to divide the datasets by "Day". Specifically, we first split datasets by "Day" granularity with an approximate splitting rate equal to 2/3-1/6-1/6 for training, validation, and test sets. Within each data split, we further aggregate the events that happen within the same time interval according to the given granularity (second, minute, hour or day). This splitting mechanism effectively prevents data leakage issues, as the datasets for different time granularities are split at the identical timestamp. Simultaneously, the semantics of the edges remain unchanged, enabling fair cross-granularity comparisons. Table~\ref{tab:data_split} outlines the specific dataset splits.

\vspace{-1mm}
\subsection{Evaluation Metrics}
\vspace{-1mm}
In order to provide a fair comparison, we employ the same evaluation metrics (AU-ROC and AP) as the benchmark paper~\cite{poursafaei_towards_2022} to assess the performance of the models. AU-ROC~\cite{hosmer2013area} (Area Under the Receiver Operating Characteristic Curve) summarizes the ROC curve into a single number that describes the performance of a model for multiple thresholds at the same time; while AP (Average Precision) is calculated as the weighted mean of precisions at each threshold. Both metrics range from 0 to 1, with higher scores indicating better performance. They are widely utilized in dynamic link prediction tasks due to their robustness against imbalanced data distribution and their adaptability across various classification thresholds.

\vspace{-1mm}
\subsection{The Baseline and DGNNs}
\vspace{-1mm}

As a natural extension of~\cite{poursafaei_towards_2022}, we retain the EdgeBank~\cite{poursafaei_towards_2022} as our baseline and select three Dynamic Graph Neural Networks (DGNNs) for comparison, namely JODIE~\cite{kumar2019predicting}, DyRep~\cite{trivedi2018dyrep}, and TGN~\cite{rossi2020temporal}. Note that EdgeBank is a non-parametric approach purely based on memorization, and other approaches are specifically designed to handle dynamic graphs and have achieved state-of-the-art performance in link prediction at the time of their release.


\paragraph{EdgeBank~\cite{poursafaei_towards_2022}} stores observed edges in a dictionary and updates its memory at each timestamp. It predicts a test edge as positive if this edge was seen before and negative otherwise. It has two variants: $\mathbf{EdgeBank_{\infty}}$ stores all observed edges in memory and is more adept at identifying rare edges, while $\mathbf{EdgeBank_{tw}}$ only remembers edges from the short-term past. We set the time window to be the same size as the validation set in our experiments. EdgeBank is a simple but strong baseline in the link prediction task; the recent work~\cite{poursafaei_towards_2022} argues that any valid DGNN method should outperform EdgeBank. 

\paragraph{JODIE~\cite{kumar2019predicting}} tackles graph dynamics by predicting the embedding trajectory in temporal interaction networks. It captures the time-evolving nature of the interactions by jointly learning the embeddings and their projections in time, allowing the model to adapt to changes in the graph structure. By incorporating a recurrent architecture with a dedicated update mechanism, JODIE can efficiently learn and update node embeddings, making it highly effective in capturing temporal dependencies and predicting future interactions in dynamic graphs.

\paragraph{DyRep~\cite{trivedi2018dyrep}} introduces a novel framework that captures both topological and temporal dependencies by employing a graph attention mechanism for structural information and a point process-based approach for temporal dynamics. DyRep can learn node embeddings that effectively represent both the graph structure and the temporal evolution, allowing the model to generalise well on various dynamic graph tasks.

\paragraph{TGN~\cite{rossi2020temporal}} provides a generic, scalable and efficient framework to model dynamic graphs. It addresses the challenges of dynamic graphs by incorporating memory modules and a message-passing mechanism that can capture both structural and temporal information. The memory modules store historical node embeddings, while the message-passing mechanism allows nodes to exchange information with their neighbours. TGN can efficiently handle dynamic graphs by employing a combination of attention mechanisms and temporal aggregators to learn node representations that capture the local and global temporal dependencies.

\subsection{Negative Sampling}

\looseness-1
In this research, we employ the three negative sampling strategies proposed in~\cite{poursafaei_towards_2022} to conduct a comprehensive and robust evaluation of various methods across multiple time granularities. Specifically, \textbf{Random Negative Sampling (RandNS)} involves selecting negative edges at random from any node pairs within the graph. \textbf{Historical Negative Sampling (HistNS)} chooses negative edges that are previously observed but do not recur in the current testing phase, with the aim of assessing the model's ability to predict recurring edges. \textbf{Inductive Negative Sampling (InduNS)} evaluates the model's capacity to handle reoccurrence patterns of unseen edges by constructing edges not observed during training. The latter two sampling techniques address the inherent limitations of random sampling and challenge DGNNs in more stringent settings. Furthermore, these strategies do not interfere with time granularity, thus allowing us to maintain focus on this crucial aspect while still benefiting from the robustness offered by the diverse negative sampling approaches.

\vspace{-2mm}
\section{Experiment Results \& Discussion}
\vspace{-1mm}
We conducted extensive experiments to evaluate the performance and robustness of various models for the dynamic link prediction task under different settings. Specifically, we trained baseline models and selected DGNN models on each dataset with four predetermined time granularities, resulting in a total of 140 models (5 methods × 7 datasets × 4 time granularities). To differentiate the models by their time granularity, we added a suffix (-s, -m, -h, or -d) to the model names, indicating training at the second, minute, hour, and day time granularities, respectively. We then evaluated the trained models across different time granularities for each of the three negative sampling settings. For example, a TGN model trained on the Wikipedia dataset under the "second" time granularity (TGN-s) would be compared with two baselines and two other competitive models (JODIE-s \& DyRep-s) trained under the same setting, as well as with other TGNs (TGN-m/h/d) trained at different time granularities. Each comparison was conducted in three negative sampling settings to obtain a comprehensive view and a comparative ranking of the given model.

We tested models trained on fine-grained time granularities (referred to as \textbf{"fine models"}, while models trained on coarse-grained time granularities are called \textbf{"coarse models"} in the following text) on coarse-grained test sets to examine the significance of time granularity in message passing and model training. We anticipated that fine models should achieve at least the same performance as coarse models when tested on the corresponding coarse time granularity used to train the coarse models. We also evaluated coarse models on fine-grained test sets to investigate their robustness to changes in time granularity. We expected that the performance of coarse models would be limited by the corresponding fine models' performance due to the inevitable and irrecoverable information loss. Simultaneously, we calculated the relative gain or loss in performance when a model was tested on a time granularity different from the one used in its training.

All experiments were conducted using the same training configuration and hyperparameters to maintain consistency and comparability across the results; for more details, refer to Appendix~\ref{hyperparameter}. The experiments were performed on Google Colab utilizing an A100 GPU, and the reported outcomes represent the average results obtained over three runs.

\begin{table}[!tbp]
\centering
\caption{Average rank of AU-ROC on dynamic link prediction for different time granularities over three negative sampling strategies. Note that the top three methods are coloured by \textbf{\color[HTML]{FFD700} First}, \textbf{\color[HTML]{8B4513} Second} and \textbf{\color[HTML]{0070C0} Third} respectively. Note that the absolute difference between any two given methods can be determined by calculating the difference in their numerical scores in Appendix~\ref{evaluation}.}
\label{ref:rank}
\resizebox{\textwidth}{!}{
\begin{tabular}{l|ccc|ccc|ccc|ccc}
\toprule
\textbf{Granularity} & \multicolumn{3}{c|}{\textbf{Second}}  & \multicolumn{3}{c|}{\textbf{Minute}}  & \multicolumn{3}{c|}{\textbf{Hour}}    & \multicolumn{3}{c}{\textbf{Day}}     \\
\midrule
\textbf{NS}          & Rand       & Hist       & Indu      & Rand       & Hist       & Indu      & Rand       & Hist       & Indu      & Rand       & Hist       & Indu      \\
\midrule
JODIE-s              & \textbf{\color[HTML]{0070C0} 3} & 11         & 9          & \textbf{\color[HTML]{8B4513} 2} & 11         & 9          & \textbf{\color[HTML]{0070C0} 3} & 11         & 10         & 4          & 12         & 11         \\
DyRep-s              & 12         & 7          & 6          & 11         & 7          & 6          & 14         & 7          & 6          & 14         & 7          & 5          \\
TGN-s                & 6          & \textbf{\color[HTML]{8B4513} 2} & \textbf{\color[HTML]{FFD700} 1} & 5          & \textbf{\color[HTML]{FFD700} 1} & \textbf{\color[HTML]{FFD700} 1} & 6          & 5          & \textbf{\color[HTML]{0070C0} 3} & 9          & 5          & 4          \\
JODIE-m              & \textbf{\color[HTML]{FFD700} 1} & 12         & 14         & \textbf{\color[HTML]{FFD700} 1} & 12         & 12         & \textbf{\color[HTML]{8B4513} 2} & 13         & 14         & \textbf{\color[HTML]{FFD700} 1} & 13         & 12         \\
DyRep-m              & 13         & 9          & 8          & 12         & 8          & 7          & 13         & 8          & 7          & 12         & 9          & 8          \\
TGN-m                & 5          & \textbf{\color[HTML]{FFD700} 1} & \textbf{\color[HTML]{8B4513} 2} & 4          & \textbf{\color[HTML]{8B4513} 2} & \textbf{\color[HTML]{8B4513} 2} & 7          & 5          & 4          & 7          & 4          & \textbf{\color[HTML]{0070C0} 3} \\
JODIE-h              & \textbf{\color[HTML]{8B4513} 2} & 14         & 11         & \textbf{\color[HTML]{0070C0} 3} & 14         & 14         & \textbf{\color[HTML]{FFD700} 1} & 14         & 13         & \textbf{\color[HTML]{8B4513} 2} & 14         & 14         \\
DyRep-h              & 14         & 8          & 7          & 13         & 6          & 5          & 11         & 6          & 5          & 13         & 8          & 6          \\
TGN-h                & 8          & 5          & \textbf{\color[HTML]{0070C0} 3} & 7          & 5          & \textbf{\color[HTML]{0070C0} 3} & 5          & \textbf{\color[HTML]{FFD700} 1} & \textbf{\color[HTML]{FFD700} 1} & 6          & \textbf{\color[HTML]{0070C0} 3} & \textbf{\color[HTML]{8B4513} 2} \\
JODIE-d              & 4          & 13         & 12         & 6          & 13         & 13         & 4          & 12         & 11         & \textbf{\color[HTML]{0070C0} 3} & 11         & 13         \\
DyRep-d              & 11         & 6          & 5          & 14         & 9          & 8          & 12         & 9          & 8          & 11         & 6          & 7          \\
TGN-d                & 9          & 4          & 4          & 10         & 4          & 4          & 8          & \textbf{\color[HTML]{8B4513} 2} & \textbf{\color[HTML]{8B4513} 2} & 5          & \textbf{\color[HTML]{FFD700} 1} & \textbf{\color[HTML]{FFD700} 1} \\
$\mathbf{EdgeBank_{tw}}$           & 7          & \textbf{\color[HTML]{0070C0} 3} & 13         & 8          & \textbf{\color[HTML]{0070C0} 3} & 11         & 9          & \textbf{\color[HTML]{0070C0} 3} & 12         & 8          & \textbf{\color[HTML]{8B4513} 2} & 10         \\
$\mathbf{EdgeBank_{\infty}}$           & 10         & 10         & 10         & 9          & 10         & 10         & 10         & 10         & 9          & 10         & 10         & 9\\
\bottomrule
\end{tabular}}
\vspace{-5mm}
\end{table}

\subsection{Overall Performance of Dynamic Link Prediction}

Table~\ref{ref:rank} shows the average rankings for model performance on different granularities based on AU-ROC. The model rankings for AP are consistent with the results obtained for AU-ROC. To conduct a meticulous evaluation of models trained on diverse granularities and negative sampling strategies, we have incorporated 24 supplementary tables presenting numerical results in Appendix~\ref{evaluation}. Each table presents the numerical performance of different models under various time granularities for each selected dataset, employing a specific negative sampling technique. These tables also exhibit the corresponding variations between different runs of experiments, as measured by standard deviation \footnote{Some standard deviations reported in the Appendix~\ref{evaluation} are rounded to 0.000 due to their extremely small magnitudes ($<1e-3$), aiming to maintain a neat and uniform format across all tables.}. The performances of the models are ranked, and the average rankings are consolidated in Table~\ref{ref:rank}. 

We also notice that JODIE-x (including "JODIE-s", "JODIE-m", "JODIE-h", and "JODIE-d") models surpass other models across all granularities when the test set is randomly sampled. However, under alternative negative sampling strategies, their performance declines significantly, ranking near the bottom. DyRep-x models maintain consistent performance across all granularities. Although they do not have remarkable performance in any specific dataset, DyRep-x slightly outperforms JODIE-x in HistNS and InduNS settings. TGN-x demonstrates stable, robust performance across all datasets in any negative sampling setting and exhibits a substantial lead in challenging test environments. EdgeBank remains a competitive baseline in our experiments, particularly in the HistNS setting, where EdgeBank with a fixed time window, secures the top position. 

\vspace{-1mm}
\subsection{Cross-granularity Comparison}
\vspace{-1mm}
It is noteworthy that certain coarse models achieve comparable or even marginally superior performance than their fine counterparts when evaluated on fine-grained test sets. For instance, JODIE-m and JODIE-h achieve similar performance to JODIE-s when tested on the "second" time granularity for all datasets. Likewise, TGN-s and TGN-m demonstrate no significant performance discrepancies in both the "second" and "minute" granularities. In fact, TGN-m yields marginally higher scores on the MOOC and LastFM datasets. These counter-intuitive instances indicate that training on the finest granularity may not always be the optimal choice, as fine-grained timestamps may not provide useful information for the underlying task and could introduce additional noise during training.

Another important observation is that the distance between the target time granularity and the current time granularity used in training has a considerable impact on model performance. For example, when handling predictions in the "day" time interval, a model trained in "day" or "hour" granularities typically outperforms models trained at finer granularities ("second" and "minute"), despite the coarse model experiencing information loss.

\vspace{-1mm}
\subsection{Model Design and Performance}
\vspace{-1mm}
To investigate the reasons behind performance gaps among the selected DGNNs, we explored their internal architectures for memorization and message passing. TGN selectively stores the memory for previously encountered edges through the forget gates in Gated Recurrent Units (GRUs), allowing it to remember crucial interactions over long distances while mitigating memory strain due to repeated updates. In contrast, JODIE and DyRep employ vanilla RNNs to memorize previous edges, with each update potentially diluting their memory. This explains why TGN-x models consistently achieve strong and robust performance across various time granularities under different testing environments. DyRep-x models consistently rank lower in performance within the scope of our experiment. One plausible explanation for this is that when computing new messages for newly occurring events, DyRep disregards the interactions between the event and the destination node, while the other two approaches thoroughly consider both the involved nodes' embeddings, interactions and previous memory. To summarize, although JODIE and DyRep can be conceptualized as special cases of TGN, our experimental insights underscore that TGN offers a highly adaptable framework for addressing a wide range of dynamic network-related tasks across various domains.

\begin{figure}[!t]
  \centering
  \subfigure[]{
    \includegraphics[width=0.3\textwidth]{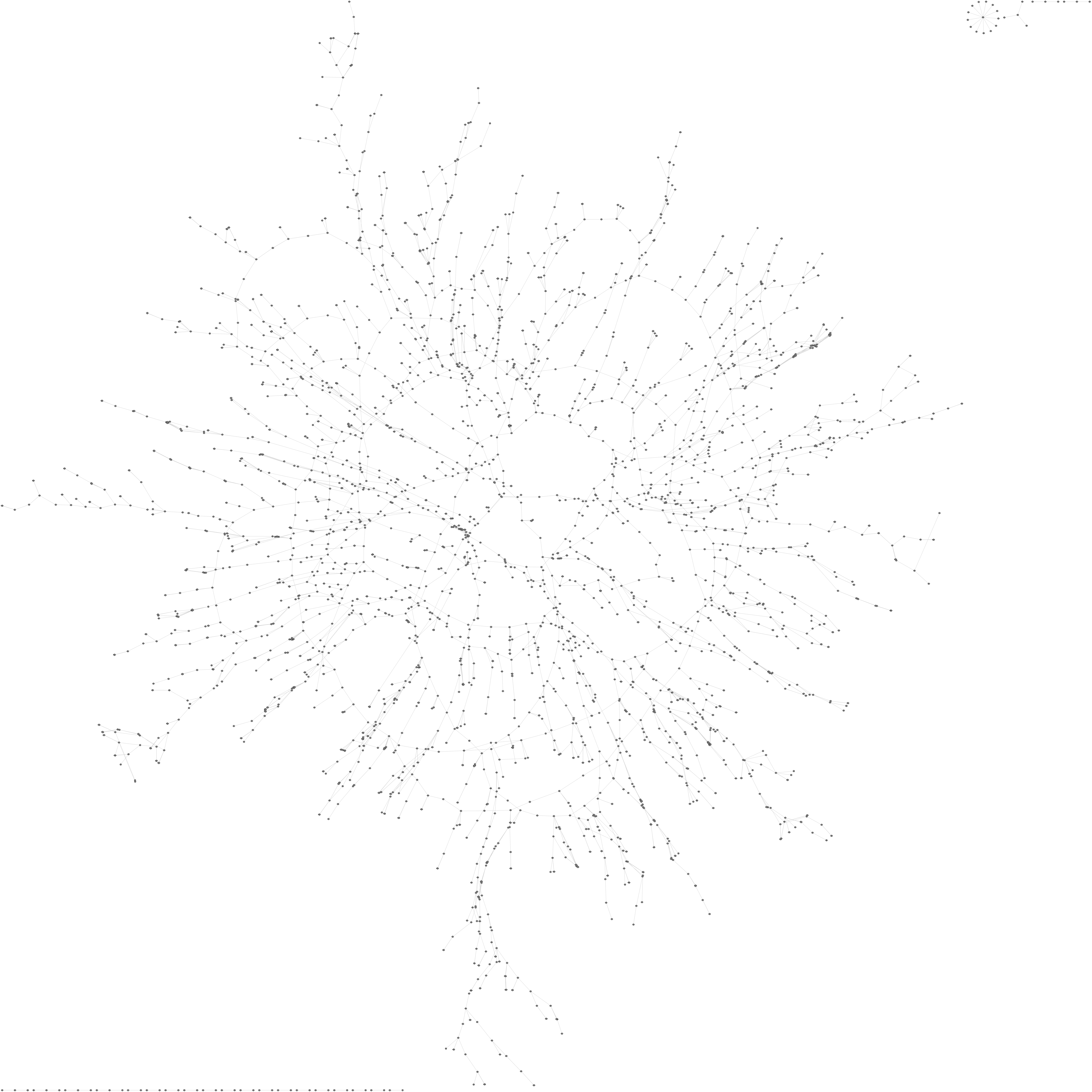}
  }
  \subfigure[]{
    \includegraphics[width=0.3\textwidth]{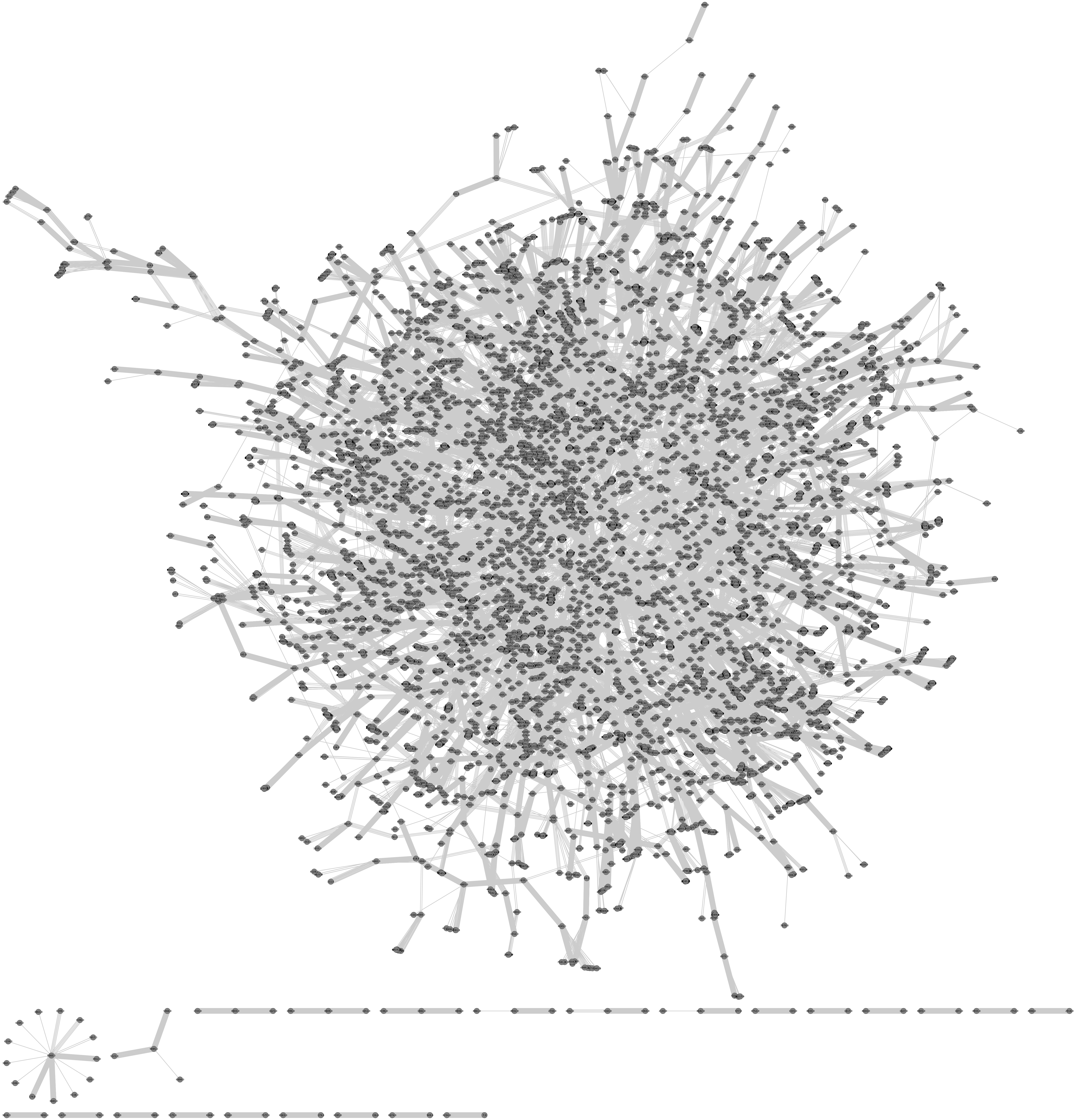}
  }
  \subfigure[]{
    \includegraphics[width=0.3\textwidth]{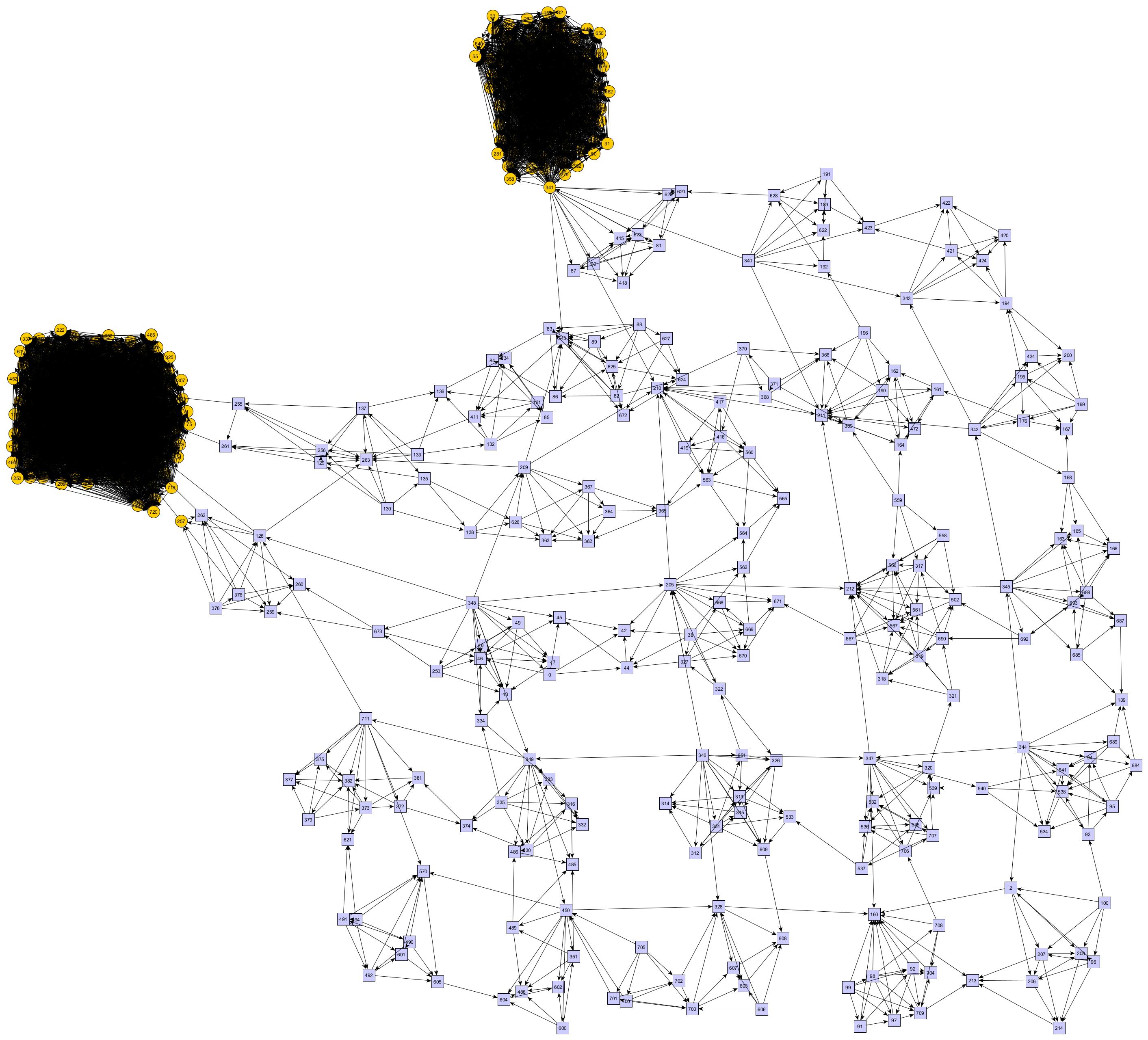}
  }
  \looseness-1
  \caption{An example of "hairball" graph due to repetitive edge additions and aggregation. (a) Original Wikipedia graph used in our experiment (no edge repetition); (b) The "hairball" visualisation of the Wikipedia graph under our edge aggregation method; (c) A synthetic example of a globally sparse but locally dense graph, containing multiple "black holes". (a) and (b) are visualised using the Backbone layout~\cite{nocaj2015untangling} in Visone~\cite{Visone} without edge sparsification. The width of the edge indicates the number of communications between two designated edges. (c) is visualised using the Organic layout~\cite{organiclayout} in yEd~\cite{yed}.}
  \label{hairball}
  \vspace{-7mm}
\end{figure}

All selected DGNNs update their node embeddings and memories for each batch during training. Given that no edge merging or de-duplication operations occur, the total number of training instances remains consistent for all models trained at different time granularities (refer to Section~\ref{notation} for the edge aggregation method used in our experiments). This suggests that all models share the same complexity, and differences in training time can be disregarded. In our experiments, no trade-off exists between computational costs and model performance. Instead, we purely focus on evaluating whether the model can effectively address the problem at finer time granularities when the corresponding time information is eliminated. However, one could argue that these models, particularly the TGN-x models, capture superficial patterns from training instances, akin to "short-cut" learning observed in other research domains. As the total training instances remain unchanged, the model may attempt to recover lost timestamp information from the input sequence, leveraging the learned ordering information to predict link existence at finer granularities.

\vspace{-4mm}
\subsection{Drawbacks of Benchmark Datasets}
\vspace{-2mm}
\looseness-1
In all cases except for the RandNS setting, where test samples are too simplistic to differentiate between various methods, we observe a model's performance ranking on one dataset to be approximately consistent with its rankings on other datasets. This observation can be reasonably attributed to the fact that all selected datasets are large, complex networks extracted from the real world. Despite classifying them into different domains based on their associated meanings, these scale-free networks, which exhibit a power-law degree distribution, share many topological characteristics. It limits our capabilities in evaluating the model's robustness over varied time granularities across different network structures.

In our experiments, we consider dynamic graphs without edge or node deletion events and with edges added sequentially, resulting in a high prevalence of duplicated edges. Table 1 illustrates that, in extreme cases, over 90\% of the edges are duplicated. In addition, our edge aggregation method ensures that each snapshot of the dynamic graph contains all edges present in the previous steps. Consequently, although the original graph in our datasets is sparse, as demonstrated in Fig.~\ref{hairball}(a), the addition of numerous repeated edges transforms it into a complex "hairball" graph, shown in Fig.~\ref{hairball}(b). Hairball graphs~\cite{PhysRevE.93.012304}, characterized by overlapping and entangled vertices and edges, hinder the identification of meaningful patterns or structures and impede graph analysis.

Under our assumptions and simplifications, the aggregated view of the dynamic graph over time becomes a globally sparse but locally dense graph, containing multiple "black holes" due to repeated edge addition, as illustrated in Fig.~\ref{hairball}(c). The majority of the edges reside in these "black holes," leading to biased link prediction. This could partially explain the decent performance of all selected models in the RandNS setting. These models may not require any understanding of temporal information; instead, for any sampled edge within a "black hole," the models can confidently predict it as Positive, otherwise as Negative. The InduNS approach mitigates this issue to some extent, resulting in a noticeable decrease in model performance. We question whether the selected models can truly capture temporal information under our edge aggregation method and RandNS setting, let alone manage time granularities. More experiments are needed to verify our conjecture. 

\vspace{-1mm}
Our experimental results highlight significant differences in model performance with the benchmark paper~\cite{poursafaei_towards_2022}, particularly for HistNS and InduNS settings. Given that each method converges within 20 epochs and the performance standard deviations among different runs are minimal, we attribute these discrepancies in model performance primarily to different data splitting mechanisms and unbalanced distribution of edge occurrence over time shown in Table~\ref{tab:data_split}.

\vspace{-3mm}
\paragraph{Limitations}
This research inherits several limitations from previous relevant study~\cite{poursafaei_towards_2022}. Firstly, all datasets were partitioned at a single point, which is a common practice but potentially weakens the influence of temporal information on the underlying task. Secondly, all models were evaluated in a transductive setting where all nodes were seen during training. Thirdly, our evaluation only focused on a narrow aspect of dynamic network analysis, specifically dynamic link prediction, thereby limiting the scope of our findings to this particular area. Our research also uncovered new limitations. We observed that all selected DGNNs share a similar training pipeline, and all datasets are structurally analogous. These factors restricted our ability to conduct a more comprehensive analysis of model performance. Therefore, our conclusions are limited to a specific type of graph and a specific category of GNNs. Another huge barrier lies in the high demand for computation resources. The number of experiments grows four times compared with the previous work, and each model requires an enormous amount of time for training and evaluation.  

\vspace{-2mm}
\section{Conclusion \& Future Work}
\vspace{-2mm}

In this study, we explored the influence of time granularity on dynamic link prediction tasks. Our methodology encompassed comprehensive experiments using EdgeBank as the baseline model, along with three dynamic graph neural networks (DGNNs) - JODIE, DyRep, and TGN - trained across seven datasets at four distinct time granularities (second, minute, hour, and day). The evaluation included three negative sampling strategies and extensive cross-granularity testing to assess the models' robustness against varying time information. Our findings revealed that TGN consistently outperformed other models across different settings, attributable to its sophisticated memorization mechanism and message-passing pipeline. Notably, we observed that models with coarser granularity sometimes matched or even exceeded the performance of finer-grained models, suggesting that fine-grained time information is not always beneficial and might introduce noise.

\vspace{-1mm}
Our research marks a foundational step in understanding the role of time granularity in dynamic graph analysis, and there is significant potential for further investigation. Recent advancements in sophisticated models, including TREND~\cite{wen2022trend}, NAT~\cite{luo2022neighborhood}, and CAWs~\cite{wang2021inductive}, present opportunities for future exploration. Extending our experimental framework to different types of dynamic graphs across various domains and incorporating diverse model architectures would enrich our understanding of this field. Furthermore, conducting node-level or graph-level experiments could offer a more holistic view of time granularity's impact on dynamic graph analysis.

\vspace{-1mm}
Some innovative modifications of our pipeline, such as avoiding fixed point split for datasets~\cite{roland2022}, modifying edge aggregation methods, or eliminating negative sampling for graphs with smaller scales~\cite{roland2022}, are promising directions for future research. Another intriguing possibility is to aggregate events within coarser time intervals, remove duplicate edges, and record occurrence frequencies or probabilities. This approach could reduce computational demands and enable more sophisticated prediction tasks, such as estimating the number of events (e.g., number of flights between cities) within a specified timeframe. Finally, considering the uneven distribution of edge occurrences, there is scope for designing models that aggregate events in a data-driven manner. Employing learnable time granularity, as opposed to deterministic aggregation at pre-set timestamps, could lead to more nuanced and effective dynamic graph analyses.

\section{Acknowledgements}
The authors would like to thank Prof. Pietro Lio, Dr. Petar Veličković, Dr. Farimah Poursafaei, and Shenyang (Andy) Huang for their insightful discussions regarding this work and valuable suggestions on polishing earlier versions of the manuscript.

\bibliography{reference}

\begin{thebibliography}{10}

\bibitem{wu2022graph}
Shiwen Wu, Fei Sun, Wentao Zhang, Xu~Xie, and Bin Cui.
\newblock Graph neural networks in recommender systems: a survey.
\newblock {\em ACM Computing Surveys}, 55(5):1--37, 2022.

\bibitem{gao2023survey}
Chen Gao, Yu~Zheng, Nian Li, Yinfeng Li, Yingrong Qin, Jinghua Piao, Yuhan Quan, Jianxin Chang, Depeng Jin, Xiangnan He, et~al.
\newblock A survey of graph neural networks for recommender systems: Challenges, methods, and directions.
\newblock {\em ACM Transactions on Recommender Systems}, 1(1):1--51, 2023.

\bibitem{yang2022s}
Xuan Yang, Yang Yang, Jintao Su, Yifei Sun, Zhongyao Wang, Shen Fan, Jun Zhang, and Jingmin Chen.
\newblock Who's next: Rising star prediction via diffusion of user interest in social networks.
\newblock {\em IEEE Transactions on Knowledge and Data Engineering}, 2022.

\bibitem{zhao2019t}
Ling Zhao, Yujiao Song, Chao Zhang, Yu~Liu, Pu~Wang, Tao Lin, Min Deng, and Haifeng Li.
\newblock T-gcn: A temporal graph convolutional network for traffic prediction.
\newblock {\em IEEE transactions on intelligent transportation systems}, 21(9):3848--3858, 2019.

\bibitem{zhang2019spatial}
Chenhan Zhang, JQ~James, and Yi~Liu.
\newblock Spatial-temporal graph attention networks: A deep learning approach for traffic forecasting.
\newblock {\em IEEE Access}, 7:166246--166256, 2019.

\bibitem{zhou2020variational}
Fan Zhou, Qing Yang, Ting Zhong, Dajiang Chen, and Ning Zhang.
\newblock Variational graph neural networks for road traffic prediction in intelligent transportation systems.
\newblock {\em IEEE Transactions on Industrial Informatics}, 17(4):2802--2812, 2020.

\bibitem{wang2022causalgnn}
Lijing Wang, Aniruddha Adiga, Jiangzhuo Chen, Adam Sadilek, Srinivasan Venkatramanan, and Madhav Marathe.
\newblock Causalgnn: Causal-based graph neural networks for spatio-temporal epidemic forecasting.
\newblock In {\em Proceedings of the AAAI Conference on Artificial Intelligence}, volume~30, pages 12191--12199, 2022.

\bibitem{adiga2022ai}
Aniruddha Adiga, Bryan Lewis, Simon Levin, Madhav~V Marathe, H~Vincent Poor, SS~Ravi, Daniel~J Rosenkrantz, Richard~E Stearns, Srinivasan Venkatramanan, Anil Vullikanti, et~al.
\newblock Ai techniques for forecasting epidemic dynamics: Theory and practice.
\newblock In {\em Artificial Intelligence in Covid-19}, pages 193--228. Springer, 2022.

\bibitem{longa_graph_2023}
Antonio Longa, Veronica Lachi, Gabriele Santin, Monica Bianchini, Bruno Lepri, Pietro Lio, Franco Scarselli, and Andrea Passerini.
\newblock Graph neural networks for temporal graphs: State of the art, open challenges, and opportunities, 2023-02-03.

\bibitem{chen2021learning}
Tailin Chen, Desen Zhou, Jian Wang, Shidong Wang, Yu~Guan, Xuming He, and Errui Ding.
\newblock Learning multi-granular spatio-temporal graph network for skeleton-based action recognition.
\newblock In {\em Proceedings of the 29th ACM international conference on multimedia}, pages 4334--4342, 2021.

\bibitem{skarding2021foundations}
Joakim Skarding, Bogdan Gabrys, and Katarzyna Musial.
\newblock Foundations and modeling of dynamic networks using dynamic graph neural networks: A survey.
\newblock {\em IEEE Access}, 9:79143--79168, 2021.

\bibitem{mao2019spatio}
Zhenyu Mao, Yi~Su, Guangquan Xu, Xueping Wang, Yu~Huang, Weihua Yue, Li~Sun, and Naixue Xiong.
\newblock Spatio-temporal deep learning method for adhd fmri classification.
\newblock {\em Information Sciences}, 499:1--11, 2019.

\bibitem{mubashar2022efficient}
Rida Mubashar, Mazhar~Javed Awan, Muhammad Ahsan, Awais Yasin, and Vishwa~Pratap Singh.
\newblock Efficient residential load forecasting using deep learning approach.
\newblock {\em International Journal of Computer Applications in Technology}, 68(3):205--214, 2022.

\bibitem{qian2021exploring}
Rui Qian, Yeqing Li, Liangzhe Yuan, Boqing Gong, Ting Liu, Matthew Brown, Serge Belongie, Ming-Hsuan Yang, Hartwig Adam, and Yin Cui.
\newblock Exploring temporal granularity in self-supervised video representation learning.
\newblock {\em arXiv preprint arXiv:2112.04480}, 2021.

\bibitem{kazemi2020representation}
Seyed~Mehran Kazemi, Rishab Goel, Kshitij Jain, Ivan Kobyzev, Akshay Sethi, Peter Forsyth, and Pascal Poupart.
\newblock Representation learning for dynamic graphs: A survey.
\newblock {\em The Journal of Machine Learning Research}, 21(1):2648--2720, 2020.

\bibitem{su2020transferable}
Haisheng Su, Xu~Zhao, Tianwei Lin, Shuming Liu, and Zhilan Hu.
\newblock Transferable knowledge-based multi-granularity fusion network for weakly supervised temporal action detection.
\newblock {\em IEEE Transactions on Multimedia}, 23:1503--1515, 2020.

\bibitem{su2021transferable}
Haisheng Su, Peiqin Zhuang, Yukun Li, Dongliang Wang, Weihao Gan, Wei Wu, and Yu~Qiao.
\newblock Transferable knowledge-based multi-granularity aggregation network for temporal action localization: Submission to activitynet challenge 2021.
\newblock {\em arXiv preprint arXiv:2107.12618}, 2021.

\bibitem{rossi2020temporal}
Emanuele Rossi, Ben Chamberlain, Fabrizio Frasca, Davide Eynard, Federico Monti, and Michael Bronstein.
\newblock Temporal graph networks for deep learning on dynamic graphs.
\newblock {\em arXiv preprint arXiv:2006.10637}, 2020.

\bibitem{kumar2019predicting}
Srijan Kumar, Xikun Zhang, and Jure Leskovec.
\newblock Predicting dynamic embedding trajectory in temporal interaction networks.
\newblock In {\em Proceedings of the 25th ACM SIGKDD international conference on knowledge discovery \& data mining}, pages 1269--1278, 2019.

\bibitem{wang2021inductive}
Yanbang Wang, Yen-Yu Chang, Yunyu Liu, Jure Leskovec, and Pan Li.
\newblock Inductive representation learning in temporal networks via causal anonymous walks.
\newblock {\em arXiv preprint arXiv:2101.05974}, 2021.

\bibitem{poursafaei_towards_2022}
Farimah Poursafaei, Shenyang Huang, Kellin Pelrine, and Reihaneh Rabbany.
\newblock Towards better evaluation for dynamic link prediction, 2022-09-11.

\bibitem{kazemi_representation_2020}
Seyed~Mehran Kazemi, Rishab Goel, Kshitij Jain, Ivan Kobyzev, Akshay Sethi, Peter Forsyth, and Pascal Poupart.
\newblock Representation learning for dynamic graphs: A survey, 2020-04-27.

\bibitem{skarding_foundations_2021}
Joakim Skarding, Bogdan Gabrys, and Katarzyna Musial.
\newblock Foundations and modelling of dynamic networks using dynamic graph neural networks: A survey.
\newblock {\em {IEEE} Access}, 9:79143--79168, 2021.

\bibitem{ma2018streaming}
Yao Ma, Ziyi Guo, Zhaochun Ren, Eric Zhao, Jiliang Tang, and Dawei Yin.
\newblock Streaming graph neural networks, 2018.

\bibitem{pareja2019evolvegcn}
Aldo Pareja, Giacomo Domeniconi, Jie Chen, Tengfei Ma, Toyotaro Suzumura, Hiroki Kanezashi, Tim Kaler, Tao~B. Schardl, and Charles~E. Leiserson.
\newblock Evolvegcn: Evolving graph convolutional networks for dynamic graphs, 2019.

\bibitem{wang2022inductive}
Yanbang Wang, Yen-Yu Chang, Yunyu Liu, Jure Leskovec, and Pan Li.
\newblock Inductive representation learning in temporal networks via causal anonymous walks, 2022.

\bibitem{fathy_temporalgat_2020}
Ahmed Fathy and Kan Li.
\newblock {TemporalGAT}: Attention-based dynamic graph representation learning.
\newblock In Hady~W. Lauw, Raymond Chi-Wing Wong, Alexandros Ntoulas, Ee-Peng Lim, See-Kiong Ng, and Sinno~Jialin Pan, editors, {\em Advances in Knowledge Discovery and Data Mining}, Lecture Notes in Computer Science, pages 413--423. Springer International Publishing, 2020.

\bibitem{sankar2019dynamic}
Aravind Sankar, Yanhong Wu, Liang Gou, Wei Zhang, and Hao Yang.
\newblock Dynamic graph representation learning via self-attention networks, 2019.

\bibitem{sankar_dysat_2020}
Aravind Sankar, Yanhong Wu, Liang Gou, Wei Zhang, and Hao Yang.
\newblock {DySAT}: Deep neural representation learning on dynamic graphs via self-attention networks.
\newblock In {\em Proceedings of the 13th International Conference on Web Search and Data Mining}, {WSDM} '20, pages 519--527. Association for Computing Machinery, 2020-01-22.

\bibitem{cong2023dyformer}
Weilin Cong, Yanhong Wu, Yuandong Tian, Mengting Gu, Yinglong Xia, Chun cheng Jason~Chen, and Mehrdad Mahdavi.
\newblock Dyformer: A scalable dynamic graph transformer with provable benefits on generalization ability, 2023.

\bibitem{cui2021dygcn}
Zeyu Cui, Zekun Li, Shu Wu, Xiaoyu Zhang, Qiang Liu, Liang Wang, and Mengmeng Ai.
\newblock Dygcn: Dynamic graph embedding with graph convolutional network, 2021.

\bibitem{EuzenatMontanari2005}
Jérôme Euzenat and Angelo Montanari.
\newblock {\em Time Granularity}, pages 59--118.
\newblock Foundations of Artificial Intelligence. Elsevier, 2005.
\newblock ffhal-00922282.

\bibitem{huang2023temporal}
Shenyang Huang, Farimah Poursafaei, Jacob Danovitch, Matthias Fey, Weihua Hu, Emanuele Rossi, Jure Leskovec, Michael Bronstein, Guillaume Rabusseau, and Reihaneh Rabbany.
\newblock Temporal graph benchmark for machine learning on temporal graphs.
\newblock {\em arXiv preprint arXiv:2307.01026}, 2023.

\bibitem{holme_temporal_2012}
Petter Holme and Jari Saramäki.
\newblock Temporal networks.
\newblock {\em Physics Reports}, 519(3):97--125, 2012-10.

\bibitem{hutchison_expressivity_2013}
Arnaud Casteigts, Paola Flocchini, Emmanuel Godard, Nicola Santoro, and Masafumi Yamashita.
\newblock Expressivity of time-varying graphs.
\newblock In Leszek Gąsieniec and Frank Wolter, editors, {\em Fundamentals of Computation Theory}, volume 8070, pages 95--106. Springer Berlin Heidelberg, 2013.
\newblock Series Title: Lecture Notes in Computer Science.

\bibitem{10.1145/3184558.3191526}
Giang~Hoang Nguyen, John~Boaz Lee, Ryan~A. Rossi, Nesreen~K. Ahmed, Eunyee Koh, and Sungchul Kim.
\newblock Continuous-time dynamic network embeddings.
\newblock In {\em Companion Proceedings of the The Web Conference 2018}, WWW '18, page 969–976, Republic and Canton of Geneva, CHE, 2018. International World Wide Web Conferences Steering Committee.

\bibitem{kazemi2019time2vec}
Seyed~Mehran Kazemi, Rishab Goel, Sepehr Eghbali, Janahan Ramanan, Jaspreet Sahota, Sanjay Thakur, Stella Wu, Cathal Smyth, Pascal Poupart, and Marcus Brubaker.
\newblock Time2vec: Learning a vector representation of time, 2019.

\bibitem{xu2019selfattention}
Da~Xu, Chuanwei Ruan, Sushant Kumar, Evren Korpeoglu, and Kannan Achan.
\newblock Self-attention with functional time representation learning, 2019.

\bibitem{hosmer2013area}
DW~Hosmer, S~Lemeshow, and RX~Sturdivant.
\newblock Area under the receiver operating characteristic curve.
\newblock {\em Applied Logistic Regression. Third ed: Wiley}, pages 173--182, 2013.

\bibitem{trivedi2018dyrep}
Rakshit Trivedi, Mehrdad Farajtabar, Prasenjeet Biswal, and Hongyuan Zha.
\newblock Dyrep: Learning representations over dynamic graphs.
\newblock In {\em International Conference on Learning Representations}, 2019.

\bibitem{nocaj2015untangling}
Arlind Nocaj, Mark Ortmann, and Ulrik Brandes.
\newblock Untangling the hairballs of multi-centered, small-world online social media networks.
\newblock {\em Journal of Graph Algorithms and Applications: JGAA}, 19(2):595--618, 2015.

\bibitem{Visone}
Visone.
\newblock \url{https://visone.ethz.ch/}, 2011.
\newblock Accessed on: March 23, 2023.

\bibitem{organiclayout}
Cambridge Intelligence.
\newblock Introducing the powerful organic graph layout.
\newblock \url{https://cambridge-intelligence.com/introducing-the-powerful-organic-graph-layout/}, 2021.
\newblock [Accessed: March 24, 2023].

\bibitem{yed}
yed graph editor.
\newblock \url{https://www.yworks.com/products/yed}, 2023.
\newblock Accessed on: March 23, 2023.

\bibitem{PhysRevE.93.012304}
Navid Dianati.
\newblock Unwinding the hairball graph: Pruning algorithms for weighted complex networks.
\newblock {\em Phys. Rev. E}, 93:012304, Jan 2016.

\bibitem{wen2022trend}
Zhihao Wen and Yuan Fang.
\newblock Trend: Temporal event and node dynamics for graph representation learning.
\newblock In {\em Proceedings of the ACM Web Conference 2022}, pages 1159--1169, 2022.

\bibitem{luo2022neighborhood}
Yuhong Luo and Pan Li.
\newblock Neighborhood-aware scalable temporal network representation learning.
\newblock In {\em Learning on Graphs Conference}, pages 1--1. PMLR, 2022.

\bibitem{roland2022}
Jiaxuan You, Tianyu Du, and Jure Leskovec.
\newblock Roland: Graph learning framework for dynamic graphs.
\newblock In {\em Proceedings of the 28th ACM SIGKDD Conference on Knowledge Discovery and Data Mining}, KDD '22, page 2358–2366, New York, NY, USA, 2022. Association for Computing Machinery.

\bibitem{shetty2004enron}
Jitesh Shetty and Jafar Adibi.
\newblock The enron email dataset database schema and brief statistical report.
\newblock {\em Information sciences institute technical report, University of Southern California}, 4(1):120--128, 2004.

\bibitem{madan2011sensing}
Anmol Madan, Manuel Cebrian, Sai Moturu, Katayoun Farrahi, et~al.
\newblock Sensing the" health state" of a community.
\newblock {\em IEEE Pervasive Computing}, 11(4):36--45, 2011.

\bibitem{panzarasa2009patterns}
Pietro Panzarasa, Tore Opsahl, and Kathleen~M Carley.
\newblock Patterns and dynamics of users' behavior and interaction: Network analysis of an online community.
\newblock {\em Journal of the American Society for Information Science and Technology}, 60(5):911--932, 2009.

\end{thebibliography}

\clearpage

\appendix

\renewcommand\thefigure{\thesection.\arabic{figure}}
\setcounter{figure}{0}
\renewcommand\thetable{\thesection.\arabic{table}}
\setcounter{table}{0}
\setcounter{page}{1}
\setcounter{equation}{0}

\section*{\centering{Appendix for submission ``Exploring Time Granularity on Temporal Graphs for Dynamic Link Prediction in Real-world Networks''}}

\section{Reporducibility}
\subsection{Real-word Datasets}
\label{appendix:real_data}
\begin{table}[htbp]
\centering
\caption{Datasets Statistics with associated semantic meanings.}
\label{tab:dataset_sum}
\resizebox{\textwidth}{!}{
\begin{tabular}{lrrrrrrrr}
\toprule
\textbf{Dataset} & \multicolumn{1}{l}{\textbf{Domain}} & \multicolumn{1}{c}{\textbf{Node}} & \multicolumn{1}{c}{\textbf{\# Nodes}} & \multicolumn{1}{c}{\textbf{Edge}} & \multicolumn{1}{c}{\textbf{Total Edges}} & \multicolumn{1}{c}{\textbf{Unique Edges}} & \multicolumn{1}{c}{\textbf{Unique Steps}} & \multicolumn{1}{c}{\textbf{Duration}} \\
\midrule
Wikipedia~\cite{kumar2019predicting}        & Social                              & Editors \& Wiki Pages             & 9,227                                    & Editing Request                   & 157,474                                  & 18,257                                    & 152,757                                   & 1 Month                               \\
Reddit~\cite{kumar2019predicting}           & Social                              & Uers \& Posts                     & 10,984                                   & Posting Request                   & 672,447                                  & 78,516                                    & 588,918                                   & 1 Month                               \\
MOOC~\cite{kumar2019predicting}             & Interaction                         & Students \& Online Courses        & 7,144                                    & Accessing a online course         & 411,749                                  & 178,443                                   & 345,600                                   & 1 Month                             \\
LastFM~\cite{kumar2019predicting}           & Interaction                         & Users \& Songs                    & 1,980                                    & Listening a song                  & 1,293,103                                & 154,993                                   & 1,283,614                                 & 4 Years                               \\
\bottomrule
\end{tabular}}
\end{table}

\subsection{Data Distribution} \label{data_distribution}

\begin{table}[htbp]
\caption{Data distribution in a 0.7-0.15-0.15 chronological split for different time granularities.}
\label{tab:investigate}
\resizebox{\textwidth}{!}{
\begin{tabular}{crrrrrrrrrlll}
\cline{1-10}
\multicolumn{1}{c|}{\multirow{2}{*}{Granularity}} & \multicolumn{3}{c|}{Wikipedia}                                                     & \multicolumn{3}{c|}{Reddit}                                                        & \multicolumn{3}{c}{MOOC}                                                           &                            &                           &                           \\ \cline{2-10}
\multicolumn{1}{c|}{}                           & \multicolumn{1}{c}{Train} & \multicolumn{1}{c}{Val} & \multicolumn{1}{c|}{Test}    & \multicolumn{1}{c}{Train} & \multicolumn{1}{c}{Val} & \multicolumn{1}{c|}{Test}    & \multicolumn{1}{c}{Train} & \multicolumn{1}{c}{Val} & \multicolumn{1}{c}{Test}     &                            &                           &                           \\ \cline{1-10}
\multicolumn{1}{c|}{Second}                     & 110,232                   & 23,621                  & \multicolumn{1}{r|}{23,621}  & 470,713                   & 100,667                 & \multicolumn{1}{r|}{100,867} & 288,224                   & 61,762                  & 61,763                       &                            &                           &                           \\
\multicolumn{1}{c|}{Minute}                     & 110,237                   & 23,620                  & \multicolumn{1}{r|}{23,617}  & 470,722                   & 100,858                 & \multicolumn{1}{r|}{100,867} & 288,224                   & 61,770                  & 61,755                       &                            &                           &                           \\
\hline
\end{tabular}}
\end{table}

\subsection{Hyperparameters in Model Training} \label{hyperparameter}

\begin{table}[htbp]
\centering
\caption{Training Configuration \& Hyperparameter Setting for Experiments.}
\label{tab:hyper}
\begin{tabular}{lr}
\toprule
Hyperparameter                 & Value     \\ \midrule
Learning Rate                   & 0.0001     \\
Optimizer                       & Adam       \\
Batch Size                      & 200        \\
Number of Epoches               & 20         \\
Tolerance for Eearly Stopping   & 5 Epoches  \\
Dropout Rate                    & 0.1        \\
Attention Heads                 & 2          \\
Node Embeddings                 & 100        \\
Time Embeddings                 & 100        \\
Memory Dimension (TGN)          & 172        \\
Message Dimension (TGN)         & 100        \\
Number of Runs                  & 3          \\
\bottomrule
\end{tabular}
\end{table}

\clearpage
\section{Numerical Results of Cross-granularity Evaluation} \label{evaluation}

\begin{table}[htbp]
\centering
\caption{AU-ROC of dynamic link prediction on the "second" granularity data across three negative sampling strategies. Note that we report the mean AU-ROC over three runs with the standard deviations in parenthesis, and the rank is computed by averaging the ranks over all datasets.}
\resizebox{\textwidth}{!}{
\subtable[Random Sampling]{
\begin{tabular}{lcccccccc}
\toprule
Method                       & Wikipedia     & Reddit        & MOOC          & LastFM        & Enron         & Social Evo.   & UCI           & Rank \\ \midrule
JODIE-s                      & 0.992 (0.001) & 0.999 (0.000) & 0.857 (0.015) & 0.999 (0.000) & 0.930 (0.005) & 0.996 (0.000) & 0.994 (0.001) & 3    \\
DyRep-s                      & 0.681 (0.010) & 0.566 (0.000) & 0.606 (0.005) & 0.505 (0.009) & 0.643 (0.011) & 0.679 (0.053) & 0.833 (0.001) & 12   \\
TGN-s                        & 0.979 (0.001) & 0.961 (0.000) & 0.789 (0.012) & 0.694 (0.013) & 0.808 (0.019) & 0.923 (0.001) & 0.901 (0.006) & 6    \\
JODIE-m                      & 0.993 (0.000) & 0.999 (0.000) & 0.865 (0.007) & 0.999 (0.000) & 0.926 (0.002) & 0.995 (0.001) & 0.995 (0.000) & 1    \\
DyRep-m                      & 0.677 (0.011) & 0.561 (0.000) & 0.588 (0.010) & 0.495 (0.005) & 0.634 (0.019) & 0.696 (0.015) & 0.837 (0.008) & 13   \\
TGN-m                        & 0.978 (0.001) & 0.955 (0.000) & 0.805 (0.029) & 0.695 (0.016) & 0.807 (0.020) & 0.869 (0.007) & 0.903 (0.039) & 5    \\
JODIE-h                      & 0.992 (0.001) & 0.996 (0.000) & 0.845 (0.018) & 0.997 (0.000) & 0.923 (0.017) & 0.914 (0.000) & 0.995 (0.001) & 2    \\
\bottomrule
\end{tabular}
}
}

\resizebox{\textwidth}{!}{
\subtable[Historical Sampling]{
\begin{tabular}{lcccccccc}
\toprule
Method                       & Wikipedia     & Reddit        & MOOC          & LastFM        & Enron         & Social Evo.   & UCI           & Rank         \\ \midrule
JODIE-s                      & 0.360 (0.006) & 0.373 (0.007) & 0.147 (0.010) & 0.415 (0.006) & 0.443 (0.002) & 0.621 (0.020) & 0.431 (0.037) & 11           \\
DyRep-s                      & 0.386 (0.003) & 0.421 (0.000) & 0.408 (0.008) & 0.481 (0.017) & 0.590 (0.018) & 0.715 (0.064) & 0.441 (0.016) & 7            \\
TGN-s                        & 0.815 (0.014) & 0.765 (0.000) & 0.695 (0.042) & 0.589 (0.008) & 0.657 (0.003) & 0.885 (0.009) & 0.755 (0.011) & 2            \\
JODIE-m                      & 0.364 (0.002) & 0.372 (0.011) & 0.138 (0.010) & 0.410 (0.010) & 0.445 (0.002) & 0.606 (0.023) & 0.386 (0.006) & 12           \\
DyRep-m                      & 0.370 (0.014) & 0.463 (0.000) & 0.387 (0.003) & 0.464 (0.011) & 0.591 (0.011) & 0.736 (0.010) & 0.425 (0.031) & 9            \\
TGN-m                        & 0.809 (0.003) & 0.735 (0.000) & 0.714 (0.049) & 0.641 (0.031) & 0.656 (0.004) & 0.846 (0.044) & 0.714 (0.063) & 1            \\
JODIE-h                      & 0.360 (0.010) & 0.315 (0.000) & 0.119 (0.002) & 0.328 (0.001) & 0.444 (0.002) & 0.528 (0.000) & 0.383 (0.042) & 14           \\
\bottomrule
\end{tabular}
}
}

\resizebox{\textwidth}{!}{
\subtable[Inductive Sampling]{
\begin{tabular}{lcccccccc}
\toprule
Method                       & Wikipedia     & Reddit        & MOOC          & LastFM        & Enron         & Social Evo.   & UCI           & Rank \\ \midrule
JODIE-s                      & 0.324 (0.004) & 0.391 (0.005) & 0.325 (0.030) & 0.465 (0.004) & 0.502 (0.001) & 0.626 (0.020) & 0.513 (0.020) & 9    \\
DyRep-s                      & 0.607 (0.009) & 0.448 (0.000) & 0.561 (0.003) & 0.505 (0.001) & 0.586 (0.014) & 0.712 (0.053) & 0.480 (0.018) & 6    \\
TGN-s                        & 0.803 (0.008) & 0.802 (0.000) & 0.654 (0.029) & 0.557 (0.005) & 0.691 (0.009) & 0.880 (0.010) & 0.746 (0.024) & 1    \\
JODIE-m                      & 0.323 (0.002) & 0.390 (0.008) & 0.308 (0.026) & 0.463 (0.006) & 0.501 (0.002) & 0.610 (0.025) & 0.479 (0.007) & 14   \\
DyRep-m                      & 0.584 (0.034) & 0.493 (0.000) & 0.521 (0.013) & 0.499 (0.008) & 0.583 (0.017) & 0.729 (0.012) & 0.471 (0.027) & 8    \\
TGN-m                        & 0.783 (0.006) & 0.790 (0.000) & 0.632 (0.014) & 0.597 (0.025) & 0.690 (0.010) & 0.837 (0.046) & 0.741 (0.030) & 2    \\
JODIE-h                      & 0.327 (0.001) & 0.366 (0.000) & 0.248 (0.018) & 0.422 (0.001) & 0.501 (0.005) & 0.526 (0.000) & 0.480 (0.036) & 11   \\
\bottomrule
\end{tabular}
}
}

\end{table}

\begin{table}[htbp]
\centering
\caption{AU-ROC of dynamic link prediction on the "minute" granularity data across three negative sampling strategies. Note that we report the mean AU-ROC over three runs with the standard deviations in parenthesis, and the rank is computed by averaging the ranks over all datasets.}
\resizebox{\textwidth}{!}{
\subtable[Random Sampling]{
\begin{tabular}{lcccccccc}
\toprule
Method                       & Wikipedia     & Reddit        & MOOC          & LastFM        & Enron         & Social Evo.   & UCI           & Rank \\ \midrule
JODIE-s                      & 0.992 (0.001) & 0.999 (0.000) & 0.856 (0.015) & 0.999 (0.000) & 0.930 (0.005) & 0.996 (0.000) & 0.994 (0.001) & 2    \\
DyRep-s                      & 0.679 (0.009) & 0.568 (0.000) & 0.603 (0.007) & 0.508 (0.008) & 0.634 (0.015) & 0.678 (0.055) & 0.830 (0.002) & 11   \\
TGN-s                        & 0.977 (0.001) & 0.961 (0.000) & 0.752 (0.026) & 0.694 (0.013) & 0.809 (0.017) & 0.847 (0.004) & 0.893 (0.009) & 5    \\
JODIE-m                      & 0.993 (0.000) & 0.999 (0.000) & 0.864 (0.007) & 0.999 (0.000) & 0.927 (0.002) & 0.995 (0.001) & 0.995 (0.000) & 1    \\
DyRep-m                      & 0.676 (0.010) & 0.559 (0.000) & 0.587 (0.008) & 0.499 (0.004) & 0.629 (0.009) & 0.697 (0.016) & 0.837 (0.010) & 12   \\
TGN-m                        & 0.977 (0.001) & 0.955 (0.000) & 0.802 (0.027) & 0.695 (0.015) & 0.810 (0.017) & 0.875 (0.007) & 0.892 (0.034) & 4    \\
JODIE-h                      & 0.992 (0.001) & 0.996 (0.000) & 0.844 (0.020) & 0.997 (0.000) & 0.923 (0.016) & 0.909 (0.000) & 0.995 (0.001) & 3    \\
\bottomrule
\end{tabular}
}
}

\resizebox{\textwidth}{!}{
\subtable[Historical Sampling]{
\begin{tabular}{lcccccccc}
\toprule
Method                       & Wikipedia     & Reddit        & MOOC          & LastFM        & Enron         & Social Evo.   & UCI           & Rank         \\ \midrule
JODIE-s                      & 0.363 (0.007) & 0.373 (0.007) & 0.148 (0.011) & 0.415 (0.006) & 0.442 (0.001) & 0.620 (0.021) & 0.431 (0.036) & 11   \\
DyRep-s                      & 0.385 (0.005) & 0.423 (0.000) & 0.404 (0.010) & 0.480 (0.019) & 0.576 (0.024) & 0.714 (0.065) & 0.441 (0.017) & 7    \\
TGN-s                        & 0.808 (0.015) & 0.763 (0.000) & 0.685 (0.043) & 0.589 (0.007) & 0.661 (0.001) & 0.845 (0.013) & 0.745 (0.013) & 1    \\
JODIE-m                      & 0.366 (0.003) & 0.372 (0.011) & 0.139 (0.010) & 0.410 (0.011) & 0.443 (0.002) & 0.603 (0.022) & 0.386 (0.006) & 12   \\
DyRep-m                      & 0.367 (0.012) & 0.455 (0.000) & 0.387 (0.003) & 0.454 (0.007) & 0.584 (0.011) & 0.734 (0.015) & 0.423 (0.032) & 8    \\
TGN-m                        & 0.805 (0.002) & 0.732 (0.000) & 0.704 (0.052) & 0.640 (0.030) & 0.661 (0.002) & 0.883 (0.021) & 0.696 (0.062) & 2    \\
JODIE-h                      & 0.362 (0.011) & 0.316 (0.001) & 0.119 (0.002) & 0.327 (0.001) & 0.443 (0.002) & 0.526 (0.000) & 0.384 (0.041) & 14   \\
\bottomrule
\end{tabular}
}
}

\resizebox{\textwidth}{!}{
\subtable[Inductive Sampling]{
\begin{tabular}{lcccccccc}
\toprule
Method                       & Wikipedia     & Reddit        & MOOC          & LastFM        & Enron         & Social Evo.   & UCI           & Rank \\ \midrule
JODIE-s                      & 0.324 (0.004) & 0.391 (0.005) & 0.326 (0.030) & 0.466 (0.004) & 0.501 (0.001) & 0.627 (0.021) & 0.512 (0.020) & 9    \\
DyRep-s                      & 0.608 (0.008) & 0.450 (0.000) & 0.556 (0.005) & 0.510 (0.014) & 0.577 (0.017) & 0.712 (0.054) & 0.479 (0.020) & 6    \\
TGN-s                        & 0.795 (0.009) & 0.801 (0.000) & 0.631 (0.012) & 0.557 (0.005) & 0.695 (0.007) & 0.837 (0.013) & 0.734 (0.025) & 1    \\
JODIE-m                      & 0.323 (0.002) & 0.390 (0.008) & 0.308 (0.027) & 0.463 (0.006) & 0.500 (0.002) & 0.609 (0.025) & 0.479 (0.007) & 12   \\
DyRep-m                      & 0.583 (0.034) & 0.485 (0.000) & 0.520 (0.012) & 0.485 (0.006) & 0.577 (0.014) & 0.727 (0.016) & 0.470 (0.028) & 7    \\
TGN-m                        & 0.779 (0.004) & 0.789 (0.000) & 0.622 (0.013) & 0.596 (0.024) & 0.694 (0.007) & 0.875 (0.022) & 0.722 (0.030) & 2    \\
JODIE-h                      & 0.328 (0.001) & 0.366 (0.000) & 0.249 (0.018) & 0.422 (0.001) & 0.501 (0.004) & 0.525 (0.000) & 0.480 (0.035) & 14   \\
\bottomrule
\end{tabular}
}
}

\end{table}

\begin{table}[htbp]
\centering
\caption{AU-ROC of dynamic link prediction on the "hour" granularity data across three negative sampling strategies. Note that we report the mean AU-ROC over three runs with the standard deviations in parenthesis, and the rank is computed by averaging the ranks over all datasets.}
\resizebox{\textwidth}{!}{
\subtable[Random Sampling]{
\begin{tabular}{lcccccccc}
\toprule
Method                       & Wikipedia     & Reddit        & MOOC          & LastFM        & Enron         & Social Evo.   & UCI           & Rank \\ \midrule
JODIE-s                      & 0.991 (0.001) & 0.987 (0.001) & 0.765 (0.012) & 0.993 (0.000) & 0.932 (0.005) & 0.781 (0.052) & 0.994 (0.001) & 3    \\
DyRep-s                      & 0.561 (0.011) & 0.483 (0.000) & 0.585 (0.011) & 0.504 (0.007) & 0.636 (0.013) & 0.679 (0.057) & 0.829 (0.003) & 14   \\
TGN-s                        & 0.949 (0.005) & 0.909 (0.000) & 0.616 (0.012) & 0.659 (0.015) & 0.802 (0.025) & 0.677 (0.018) & 0.854 (0.005) & 6    \\
JODIE-m                      & 0.993 (0.001) & 0.986 (0.000) & 0.760 (0.002) & 0.993 (0.000) & 0.928 (0.003) & 0.783 (0.040) & 0.995 (0.000) & 2    \\
DyRep-m                      & 0.619 (0.027) & 0.531 (0.000) & 0.560 (0.033) & 0.503 (0.009) & 0.644 (0.002) & 0.687 (0.023) & 0.835 (0.013) & 13   \\
TGN-m                        & 0.953 (0.001) & 0.850 (0.000) & 0.618 (0.041) & 0.667 (0.013) & 0.802 (0.025) & 0.657 (0.045) & 0.848 (0.057) & 7    \\
JODIE-h                      & 0.992 (0.001) & 0.992 (0.000) & 0.785 (0.009) & 0.993 (0.000) & 0.926 (0.015) & 0.898 (0.000) & 0.995 (0.001) & 1    \\
\bottomrule
\end{tabular}
}
}

\resizebox{\textwidth}{!}{
\subtable[Historical Sampling]{
\begin{tabular}{lcccccccc}
\toprule
Method                       & Wikipedia     & Reddit        & MOOC          & LastFM        & Enron         & Social Evo.   & UCI           & Rank         \\ \midrule
JODIE-s                      & 0.362 (0.005) & 0.401 (0.000) & 0.133 (0.006) & 0.417 (0.005) & 0.444 (0.004) & 0.451 (0.033) & 0.431 (0.029) & 11   \\
DyRep-s                      & 0.392 (0.010) & 0.504 (0.000) & 0.461 (0.035) & 0.478 (0.021) & 0.590 (0.016) & 0.716 (0.068) & 0.437 (0.014) & 7    \\
TGN-s                        & 0.703 (0.023) & 0.689 (0.000) & 0.597 (0.049) & 0.564 (0.015) & 0.656 (0.011) & 0.750 (0.058) & 0.624 (0.019) & 5    \\
JODIE-m                      & 0.364 (0.002) & 0.401 (0.000) & 0.125 (0.004) & 0.410 (0.014) & 0.446 (0.001) & 0.441 (0.018) & 0.385 (0.014) & 13   \\
DyRep-m                      & 0.399 (0.013) & 0.384 (0.000) & 0.499 (0.082) & 0.463 (0.002) & 0.596 (0.007) & 0.738 (0.013) & 0.425 (0.037) & 8    \\
TGN-m                        & 0.717 (0.004) & 0.665 (0.000) & 0.633 (0.022) & 0.601 (0.035) & 0.656 (0.011) & 0.644 (0.068) & 0.587 (0.102) & 5    \\
JODIE-h                      & 0.360 (0.008) & 0.319 (0.000) & 0.116 (0.003) & 0.326 (0.000) & 0.446 (0.002) & 0.517 (0.000) & 0.382 (0.050) & 14   \\
\bottomrule
\end{tabular}
}
}

\resizebox{\textwidth}{!}{
\subtable[Inductive Sampling]{
\begin{tabular}{lcccccccc}
\toprule
Method                       & Wikipedia     & Reddit        & MOOC          & LastFM        & Enron         & Social Evo.   & UCI           & Rank \\ \midrule
JODIE-s                      & 0.317 (0.003) & 0.411 (0.000) & 0.261 (0.028) & 0.463 (0.004) & 0.503 (0.001) & 0.450 (0.035) & 0.510 (0.020) & 10   \\
DyRep-s                      & 0.604 (0.012) & 0.535 (0.000) & 0.530 (0.002) & 0.508 (0.005) & 0.584 (0.010) & 0.715 (0.062) & 0.478 (0.021) & 6    \\
TGN-s                        & 0.680 (0.011) & 0.675 (0.000) & 0.497 (0.018) & 0.530 (0.019) & 0.690 (0.015) & 0.744 (0.055) & 0.628 (0.033) & 3    \\
JODIE-m                      & 0.316 (0.002) & 0.411 (0.001) & 0.243 (0.012) & 0.460 (0.009) & 0.502 (0.003) & 0.439 (0.020) & 0.476 (0.007) & 14   \\
DyRep-m                      & 0.634 (0.024) & 0.425 (0.000) & 0.516 (0.011) & 0.512 (0.013) & 0.587 (0.004) & 0.729 (0.011) & 0.472 (0.030) & 7    \\
TGN-m                        & 0.681 (0.006) & 0.663 (0.000) & 0.524 (0.019) & 0.553 (0.026) & 0.690 (0.015) & 0.641 (0.066) & 0.633 (0.056) & 4    \\
JODIE-h                      & 0.321 (0.001) & 0.362 (0.001) & 0.226 (0.013) & 0.418 (0.002) & 0.502 (0.003) & 0.519 (0.000) & 0.476 (0.036) & 13   \\
\bottomrule
\end{tabular}
}
}

\end{table}

\begin{table}[htbp]
\centering
\caption{AU-ROC of dynamic link prediction on the "day" granularity data across three negative sampling strategies. Note that we report the mean AU-ROC over three runs with the standard deviations in parenthesis, and the rank is computed by averaging the ranks over all datasets.}
\resizebox{\textwidth}{!}{
\subtable[Random Sampling]{
\begin{tabular}{lcccccccc}
\toprule
Method                       & Wikipedia     & Reddit        & MOOC          & LastFM        & Enron         & Social Evo.   & UCI           & Rank \\ \midrule
JODIE-s                      & 0.954 (0.006) & 0.931 (0.014) & 0.697 (0.003) & 0.851 (0.009) & 0.908 (0.001) & 0.628 (0.057) & 0.993 (0.001) & 4    \\
DyRep-s                      & 0.582 (0.048) & 0.500 (0.000) & 0.583 (0.023) & 0.510 (0.013) & 0.624 (0.027) & 0.582 (0.023) & 0.828 (0.010) & 14   \\
TGN-s                        & 0.916 (0.006) & 0.876 (0.000) & 0.607 (0.042) & 0.631 (0.019) & 0.738 (0.037) & 0.581 (0.061) & 0.846 (0.019) & 9    \\
JODIE-m                      & 0.993 (0.002) & 0.986 (0.001) & 0.760 (0.006) & 0.993 (0.005) & 0.928 (0.006) & 0.600 (0.040) & 0.995 (0.000) & 1    \\
DyRep-m                      & 0.619 (0.036) & 0.531 (0.000) & 0.560 (0.030) & 0.503 (0.007) & 0.644 (0.010) & 0.611 (0.017) & 0.835 (0.018) & 12   \\
TGN-m                        & 0.953 (0.003) & 0.850 (0.000) & 0.618 (0.035) & 0.667 (0.011) & 0.802 (0.038) & 0.534 (0.025) & 0.848 (0.058) & 7    \\
JODIE-h                      & 0.962 (0.004) & 0.965 (0.000) & 0.700 (0.008) & 0.857 (0.000) & 0.897 (0.014) & 0.731 (0.053) & 0.994 (0.001) & 2    \\
\bottomrule
\end{tabular}
}
}

\resizebox{\textwidth}{!}{
\subtable[Historical Sampling]{
\begin{tabular}{lcccccccc}
\toprule
Method                       & Wikipedia     & Reddit        & MOOC          & LastFM        & Enron         & Social Evo.   & UCI           & Rank         \\ \midrule
JODIE-s                      & 0.361 (0.002) & 0.366 (0.008) & 0.153 (0.003) & 0.372 (0.001) & 0.432 (0.003) & 0.402 (0.043) & 0.430 (0.030) & 12   \\
DyRep-s                      & 0.408 (0.004) & 0.476 (0.000) & 0.479 (0.048) & 0.463 (0.015) & 0.580 (0.021) & 0.613 (0.034) & 0.440 (0.022) & 7    \\
TGN-s                        & 0.688 (0.014) & 0.684 (0.000) & 0.590 (0.056) & 0.533 (0.032) & 0.633 (0.010) & 0.657 (0.055) & 0.637 (0.028) & 5    \\
JODIE-m                      & 0.364 (0.003) & 0.401 (0.005) & 0.125 (0.006) & 0.410 (0.001) & 0.446 (0.001) & 0.380 (0.030) & 0.385 (0.023) & 13   \\
DyRep-m                      & 0.399 (0.013) & 0.384 (0.000) & 0.499 (0.050) & 0.463 (0.003) & 0.596 (0.008) & 0.645 (0.011) & 0.425 (0.028) & 9    \\
TGN-m                        & 0.717 (0.014) & 0.665 (0.000) & 0.633 (0.048) & 0.601 (0.036) & 0.656 (0.011) & 0.583 (0.061) & 0.587 (0.100) & 4    \\
JODIE-h                      & 0.358 (0.001) & 0.385 (0.001) & 0.142 (0.011) & 0.368 (0.000) & 0.431 (0.001) & 0.480 (0.046) & 0.387 (0.040) & 14   \\
\bottomrule
\end{tabular}
}
}

\resizebox{\textwidth}{!}{
\subtable[Inductive Sampling]{
\begin{tabular}{lcccccccc}
\toprule
Method                       & Wikipedia     & Reddit        & MOOC          & LastFM        & Enron         & Social Evo.   & UCI           & Rank \\ \midrule
JODIE-s                      & 0.310 (0.001) & 0.433 (0.009) & 0.209 (0.017) & 0.426 (0.001) & 0.490 (0.003) & 0.414 (0.040) & 0.500 (0.020) & 11   \\
DyRep-s                      & 0.566 (0.015) & 0.519 (0.000) & 0.534 (0.050) & 0.508 (0.008) & 0.584 (0.017) & 0.606 (0.026) & 0.488 (0.025) & 5    \\
TGN-s                        & 0.665 (0.009) & 0.673 (0.000) & 0.515 (0.039) & 0.508 (0.027) & 0.645 (0.030) & 0.637 (0.043) & 0.640 (0.024) & 4    \\
JODIE-m                      & 0.316 (0.003) & 0.411 (0.007) & 0.243 (0.008) & 0.460 (0.001) & 0.502 (0.001) & 0.389 (0.026) & 0.476 (0.019) & 12   \\
DyRep-m                      & 0.634 (0.010) & 0.425 (0.000) & 0.516 (0.032) & 0.512 (0.009) & 0.587 (0.014) & 0.632 (0.008) & 0.472 (0.024) & 8    \\
TGN-m                        & 0.681 (0.011) & 0.663 (0.000) & 0.524 (0.009) & 0.553 (0.009) & 0.690 (0.031) & 0.595 (0.068) & 0.633 (0.063) & 3    \\
JODIE-h                      & 0.306 (0.001) & 0.453 (0.001) & 0.191 (0.007) & 0.424 (0.001) & 0.484 (0.004) & 0.494 (0.052) & 0.472 (0.029) & 14   \\
\bottomrule
\end{tabular}
}
}

\end{table}

\begin{table}[htbp]
\centering
\caption{AP of dynamic link prediction on the "second" granularity data across three negative sampling strategies. Note that we report the mean AP over three runs with the standard deviations in parenthesis, and the rank is computed by averaging the ranks over all datasets.}
\resizebox{\textwidth}{!}{
\subtable[Random Sampling]{
\begin{tabular}{lcccccccc}
\toprule
Method                       & Wikipedia     & Reddit        & MOOC          & LastFM        & Enron         & Social Evo.   & UCI           & Rank \\ \midrule
JODIE-s                      & 0.993 (0.001) & 0.999 (0.000) & 0.840 (0.018) & 1.000 (0.000) & 0.931 (0.006) & 0.995 (0.000) & 0.993 (0.001) & 3    \\
DyRep-s                      & 0.682 (0.008) & 0.563 (0.000) & 0.614 (0.005) & 0.527 (0.003) & 0.642 (0.014) & 0.728 (0.032) & 0.832 (0.002) & 12   \\
TGN-s                        & 0.981 (0.001) & 0.963 (0.000) & 0.764 (0.014) & 0.697 (0.014) & 0.790 (0.021) & 0.902 (0.001) & 0.903 (0.005) & 6    \\
JODIE-m                      & 0.994 (0.000) & 1.000 (0.000) & 0.848 (0.010) & 0.999 (0.000) & 0.926 (0.003) & 0.995 (0.001) & 0.994 (0.001) & 1    \\
DyRep-m                      & 0.684 (0.005) & 0.565 (0.000) & 0.597 (0.008) & 0.518 (0.004) & 0.629 (0.020) & 0.679 (0.034) & 0.831 (0.006) & 13   \\
TGN-m                        & 0.980 (0.001) & 0.954 (0.000) & 0.779 (0.024) & 0.699 (0.015) & 0.789 (0.022) & 0.821 (0.008) & 0.907 (0.036) & 5    \\
JODIE-h                      & 0.994 (0.001) & 0.996 (0.000) & 0.847 (0.018) & 0.998 (0.000) & 0.925 (0.018) & 0.861 (0.000) & 0.994 (0.001) & 2    \\
\bottomrule
\end{tabular}
}
}

\resizebox{\textwidth}{!}{
\subtable[Historical Sampling]{
\begin{tabular}{lcccccccc}
\toprule
Method                       & Wikipedia     & Reddit        & MOOC          & LastFM        & Enron         & Social Evo.   & UCI           & Rank         \\ \midrule
JODIE-s                      & 0.417 (0.002) & 0.435 (0.003) & 0.336 (0.002) & 0.454 (0.003) & 0.440 (0.001) & 0.517 (0.012) & 0.457 (0.021) & 11           \\
DyRep-s                      & 0.418 (0.001) & 0.437 (0.000) & 0.427 (0.003) & 0.476 (0.011) & 0.585 (0.021) & 0.717 (0.040) & 0.440 (0.007) & 7            \\
TGN-s                        & 0.827 (0.019) & 0.746 (0.000) & 0.644 (0.043) & 0.628 (0.010) & 0.649 (0.014) & 0.916 (0.010) & 0.777 (0.013) & 2            \\
JODIE-m                      & 0.419 (0.001) & 0.434 (0.005) & 0.334 (0.003) & 0.451 (0.006) & 0.441 (0.001) & 0.508 (0.013) & 0.430 (0.005) & 12           \\
DyRep-m                      & 0.411 (0.006) & 0.457 (0.000) & 0.417 (0.002) & 0.463 (0.006) & 0.588 (0.011) & 0.698 (0.029) & 0.435 (0.016) & 9            \\
TGN-m                        & 0.804 (0.016) & 0.704 (0.000) & 0.644 (0.046) & 0.684 (0.034) & 0.649 (0.014) & 0.820 (0.060) & 0.748 (0.056) & 1            \\
JODIE-h                      & 0.416 (0.004) & 0.403 (0.000) & 0.329 (0.001) & 0.407 (0.000) & 0.441 (0.001) & 0.469 (0.000) & 0.429 (0.023) & 14           \\
\bottomrule
\end{tabular}
}
}

\resizebox{\textwidth}{!}{
\subtable[Inductive Sampling]{
\begin{tabular}{lcccccccc}
\toprule
Method                       & Wikipedia     & Reddit        & MOOC          & LastFM        & Enron         & Social Evo.   & UCI           & Rank \\ \midrule
JODIE-s                      & 0.412 (0.002) & 0.443 (0.003) & 0.444 (0.028) & 0.477 (0.002) & 0.473 (0.000) & 0.519 (0.012) & 0.501 (0.013) & 9            \\
DyRep-s                      & 0.626 (0.005) & 0.481 (0.000) & 0.599 (0.002) & 0.519 (0.002) & 0.586 (0.018) & 0.714 (0.036) & 0.491 (0.005) & 6            \\
TGN-s                        & 0.840 (0.007) & 0.833 (0.000) & 0.658 (0.027) & 0.601 (0.008) & 0.693 (0.011) & 0.910 (0.010) & 0.782 (0.025) & 1            \\
JODIE-m                      & 0.411 (0.001) & 0.443 (0.004) & 0.430 (0.014) & 0.476 (0.004) & 0.473 (0.001) & 0.510 (0.015) & 0.480 (0.006) & 14           \\
DyRep-m                      & 0.614 (0.025) & 0.509 (0.000) & 0.558 (0.015) & 0.511 (0.006) & 0.582 (0.017) & 0.691 (0.032) & 0.499 (0.017) & 8            \\
TGN-m                        & 0.807 (0.005) & 0.807 (0.000) & 0.628 (0.022) & 0.640 (0.029) & 0.692 (0.011) & 0.808 (0.061) & 0.779 (0.024) & 2            \\
JODIE-h                      & 0.414 (0.000) & 0.429 (0.000) & 0.411 (0.018) & 0.453 (0.000) & 0.473 (0.002) & 0.467 (0.000) & 0.482 (0.020) & 11           \\
\bottomrule
\end{tabular}
}
}

\end{table}

\begin{table}[htbp]
\centering
\caption{AP of dynamic link prediction on the "minute" granularity data across three negative sampling strategies. Note that we report the mean AP over three runs with the standard deviations in parenthesis, and the rank is computed by averaging the ranks over all datasets.}
\resizebox{\textwidth}{!}{
\subtable[Random Sampling]{
\begin{tabular}{lcccccccc}
\toprule
Method                       & Wikipedia     & Reddit        & MOOC          & LastFM        & Enron         & Social Evo.   & UCI           & Rank \\ \midrule
JODIE-s                      & 0.993 (0.001) & 0.999 (0.000) & 0.840 (0.018) & 1.000 (0.000) & 0.931 (0.006) & 0.995 (0.000) & 0.993 (0.001) & 2            \\
DyRep-s                      & 0.680 (0.006) & 0.565 (0.000) & 0.611 (0.007) & 0.529 (0.005) & 0.635 (0.014) & 0.731 (0.033) & 0.827 (0.003) & 11           \\
TGN-s                        & 0.979 (0.001) & 0.963 (0.000) & 0.725 (0.023) & 0.697 (0.013) & 0.790 (0.020) & 0.784 (0.011) & 0.895 (0.009) & 5            \\
JODIE-m                      & 0.994 (0.000) & 1.000 (0.000) & 0.848 (0.009) & 0.999 (0.000) & 0.928 (0.003) & 0.994 (0.001) & 0.994 (0.001) & 1            \\
DyRep-m                      & 0.683 (0.004) & 0.562 (0.000) & 0.596 (0.007) & 0.523 (0.002) & 0.628 (0.010) & 0.677 (0.032) & 0.832 (0.008) & 12           \\
TGN-m                        & 0.979 (0.001) & 0.954 (0.000) & 0.777 (0.024) & 0.697 (0.016) & 0.791 (0.020) & 0.832 (0.009) & 0.897 (0.033) & 4            \\
JODIE-h                      & 0.994 (0.001) & 0.996 (0.000) & 0.846 (0.020) & 0.998 (0.000) & 0.924 (0.017) & 0.853 (0.000) & 0.994 (0.001) & 3            \\
\bottomrule
\end{tabular}
}
}

\resizebox{\textwidth}{!}{
\subtable[Historical Sampling]{
\begin{tabular}{lcccccccc}
\toprule
Method                       & Wikipedia     & Reddit        & MOOC          & LastFM        & Enron         & Social Evo.   & UCI           & Rank         \\ \midrule
JODIE-s                      & 0.418 (0.002) & 0.435 (0.003) & 0.337 (0.002) & 0.454 (0.003) & 0.440 (0.000) & 0.516 (0.012) & 0.456 (0.021) & 11           \\
DyRep-s                      & 0.417 (0.002) & 0.438 (0.000) & 0.425 (0.004) & 0.476 (0.013) & 0.575 (0.024) & 0.719 (0.041) & 0.440 (0.007) & 7            \\
TGN-s                        & 0.821 (0.020) & 0.742 (0.000) & 0.629 (0.043) & 0.626 (0.008) & 0.651 (0.012) & 0.838 (0.017) & 0.767 (0.014) & 1            \\
JODIE-m                      & 0.420 (0.001) & 0.434 (0.005) & 0.335 (0.002) & 0.451 (0.006) & 0.440 (0.001) & 0.507 (0.013) & 0.430 (0.005) & 12           \\
DyRep-m                      & 0.410 (0.005) & 0.453 (0.000) & 0.417 (0.001) & 0.460 (0.004) & 0.584 (0.011) & 0.694 (0.027) & 0.434 (0.016) & 8            \\
TGN-m                        & 0.802 (0.020) & 0.702 (0.000) & 0.636 (0.047) & 0.683 (0.033) & 0.651 (0.012) & 0.881 (0.039) & 0.732 (0.056) & 2            \\
JODIE-h                      & 0.417 (0.004) & 0.403 (0.000) & 0.330 (0.001) & 0.407 (0.000) & 0.440 (0.001) & 0.467 (0.000) & 0.429 (0.023) & 14           \\
\bottomrule
\end{tabular}
}
}

\resizebox{\textwidth}{!}{
\subtable[Inductive Sampling]{
\begin{tabular}{lcccccccc}
\toprule
Method                       & Wikipedia     & Reddit        & MOOC          & LastFM        & Enron         & Social Evo.   & UCI           & Rank \\ \midrule
JODIE-s                      & 0.412 (0.002) & 0.443 (0.003) & 0.445 (0.028) & 0.478 (0.002) & 0.473 (0.000) & 0.520 (0.013) & 0.500 (0.013) & 9            \\
DyRep-s                      & 0.625 (0.004) & 0.483 (0.000) & 0.594 (0.003) & 0.527 (0.011) & 0.579 (0.018) & 0.716 (0.037) & 0.491 (0.005) & 6            \\
TGN-s                        & 0.833 (0.008) & 0.832 (0.000) & 0.630 (0.012) & 0.601 (0.008) & 0.695 (0.009) & 0.825 (0.019) & 0.771 (0.025) & 1            \\
JODIE-m                      & 0.411 (0.001) & 0.443 (0.004) & 0.431 (0.015) & 0.476 (0.004) & 0.472 (0.001) & 0.510 (0.015) & 0.480 (0.006) & 12           \\
DyRep-m                      & 0.611 (0.025) & 0.504 (0.000) & 0.557 (0.010) & 0.502 (0.008) & 0.580 (0.014) & 0.688 (0.029) & 0.498 (0.016) & 7            \\
TGN-m                        & 0.805 (0.000) & 0.807 (0.000) & 0.620 (0.015) & 0.640 (0.028) & 0.695 (0.010) & 0.870 (0.039) & 0.763 (0.025) & 2            \\
JODIE-h                      & 0.415 (0.001) & 0.429 (0.000) & 0.412 (0.018) & 0.453 (0.000) & 0.473 (0.002) & 0.466 (0.000) & 0.483 (0.020) & 14           \\
\bottomrule
\end{tabular}
}
}

\end{table}

\begin{table}[htbp]
\centering
\caption{AP of dynamic link prediction on the "hour" granularity data across three negative sampling strategies. Note that we report the mean AP over three runs with the standard deviations in parenthesis, and the rank is computed by averaging the ranks over all datasets.}
\resizebox{\textwidth}{!}{
\subtable[Random Sampling]{
\begin{tabular}{lcccccccc}
\toprule
Method                       & Wikipedia     & Reddit        & MOOC          & LastFM        & Enron         & Social Evo.   & UCI           & Rank \\ \midrule
JODIE-s                      & 0.992 (0.001) & 0.989 (0.001) & 0.779 (0.010) & 0.994 (0.000) & 0.933 (0.006) & 0.760 (0.042) & 0.993 (0.001) & 3            \\
DyRep-s                      & 0.573 (0.013) & 0.515 (0.000) & 0.599 (0.006) & 0.528 (0.004) & 0.636 (0.014) & 0.730 (0.037) & 0.823 (0.002) & 14           \\
TGN-s                        & 0.952 (0.005) & 0.895 (0.000) & 0.594 (0.015) & 0.660 (0.012) & 0.775 (0.031) & 0.613 (0.024) & 0.837 (0.003) & 6            \\
JODIE-m                      & 0.994 (0.000) & 0.988 (0.000) & 0.778 (0.005) & 0.994 (0.000) & 0.929 (0.004) & 0.772 (0.031) & 0.994 (0.001) & 2            \\
DyRep-m                      & 0.639 (0.018) & 0.540 (0.000) & 0.576 (0.028) & 0.525 (0.004) & 0.643 (0.004) & 0.671 (0.047) & 0.830 (0.015) & 13           \\
TGN-m                        & 0.956 (0.002) & 0.830 (0.000) & 0.602 (0.035) & 0.664 (0.015) & 0.775 (0.031) & 0.592 (0.048) & 0.830 (0.070) & 7            \\
JODIE-h                      & 0.993 (0.001) & 0.993 (0.000) & 0.802 (0.006) & 0.994 (0.000) & 0.927 (0.017) & 0.882 (0.000) & 0.994 (0.001) & 1            \\
\bottomrule
\end{tabular}
}
}

\resizebox{\textwidth}{!}{
\subtable[Historical Sampling]{
\begin{tabular}{lcccccccc}
\toprule
Method                       & Wikipedia     & Reddit        & MOOC          & LastFM        & Enron         & Social Evo.   & UCI           & Rank         \\ \midrule
JODIE-s                      & 0.421 (0.002) & 0.449 (0.000) & 0.334 (0.002) & 0.455 (0.002) & 0.441 (0.002) & 0.441 (0.013) & 0.457 (0.017) & 11           \\
DyRep-s                      & 0.424 (0.006) & 0.486 (0.000) & 0.450 (0.014) & 0.474 (0.012) & 0.583 (0.019) & 0.720 (0.047) & 0.437 (0.006) & 7            \\
TGN-s                        & 0.689 (0.034) & 0.638 (0.000) & 0.551 (0.036) & 0.588 (0.012) & 0.635 (0.025) & 0.658 (0.073) & 0.587 (0.034) & 5            \\
JODIE-m                      & 0.422 (0.001) & 0.449 (0.000) & 0.332 (0.001) & 0.451 (0.007) & 0.442 (0.000) & 0.436 (0.006) & 0.431 (0.009) & 13           \\
DyRep-m                      & 0.424 (0.006) & 0.416 (0.000) & 0.478 (0.054) & 0.465 (0.003) & 0.592 (0.002) & 0.695 (0.036) & 0.436 (0.018) & 8            \\
TGN-m                        & 0.704 (0.008) & 0.638 (0.000) & 0.571 (0.016) & 0.627 (0.036) & 0.635 (0.024) & 0.552 (0.056) & 0.572 (0.114) & 5            \\
JODIE-h                      & 0.420 (0.003) & 0.405 (0.000) & 0.330 (0.001) & 0.407 (0.000) & 0.442 (0.001) & 0.466 (0.000) & 0.430 (0.027) & 14           \\
\bottomrule
\end{tabular}
}
}

\resizebox{\textwidth}{!}{
\subtable[Inductive Sampling]{
\begin{tabular}{lcccccccc}
\toprule
Method                       & Wikipedia     & Reddit        & MOOC          & LastFM        & Enron         & Social Evo.   & UCI           & Rank \\ \midrule
JODIE-s                      & 0.412 (0.002) & 0.454 (0.000) & 0.421 (0.029) & 0.477 (0.002) & 0.474 (0.001) & 0.441 (0.013) & 0.500 (0.013) & 10           \\
DyRep-s                      & 0.606 (0.009) & 0.555 (0.000) & 0.572 (0.004) & 0.519 (0.002) & 0.584 (0.013) & 0.720 (0.045) & 0.488 (0.006) & 6            \\
TGN-s                        & 0.686 (0.020) & 0.649 (0.000) & 0.519 (0.012) & 0.560 (0.015) & 0.679 (0.019) & 0.654 (0.069) & 0.620 (0.047) & 3            \\
JODIE-m                      & 0.411 (0.001) & 0.454 (0.000) & 0.406 (0.010) & 0.475 (0.005) & 0.474 (0.001) & 0.436 (0.007) & 0.479 (0.005) & 14           \\
DyRep-m                      & 0.650 (0.017) & 0.460 (0.000) & 0.545 (0.023) & 0.524 (0.010) & 0.587 (0.008) & 0.689 (0.036) & 0.493 (0.014) & 7            \\
TGN-m                        & 0.691 (0.005) & 0.675 (0.000) & 0.539 (0.017) & 0.581 (0.024) & 0.679 (0.019) & 0.551 (0.054) & 0.625 (0.075) & 4            \\
JODIE-h                      & 0.415 (0.001) & 0.426 (0.000) & 0.403 (0.016) & 0.452 (0.001) & 0.474 (0.002) & 0.467 (0.000) & 0.481 (0.020) & 13           \\
\bottomrule
\end{tabular}
}
}
\end{table}

\begin{table}[htbp]
\centering
\caption{AP of dynamic link prediction on the "day" granularity data across three negative sampling strategies. Note that we report the mean AP over three runs with the standard deviations in parenthesis, and the rank is computed by averaging the ranks over all datasets.}
\resizebox{\textwidth}{!}{
\subtable[Random Sampling]{
\begin{tabular}{lcccccccc}
\toprule
Method                       & Wikipedia     & Reddit        & MOOC          & LastFM        & Enron         & Social Evo.   & UCI           & Rank \\ \midrule
JODIE-s                      & 0.958 (0.006) & 0.934 (0.013) & 0.717 (0.001) & 0.882 (0.007) & 0.909 (0.001) & 0.587 (0.056) & 0.991 (0.002) & 4            \\
DyRep-s                      & 0.587 (0.047) & 0.522 (0.000) & 0.592 (0.015) & 0.531 (0.004) & 0.630 (0.023) & 0.621 (0.037) & 0.814 (0.009) & 14           \\
TGN-s                        & 0.922 (0.005) & 0.847 (0.000) & 0.584 (0.034) & 0.635 (0.017) & 0.699 (0.044) & 0.532 (0.046) & 0.835 (0.026) & 9            \\
JODIE-m                      & 0.969 (0.002) & 0.936 (0.003) & 0.711 (0.007) & 0.876 (0.004) & 0.898 (0.008) & 0.556 (0.037) & 0.993 (0.001) & 1            \\
DyRep-m                      & 0.616 (0.033) & 0.546 (0.000) & 0.575 (0.025) & 0.526 (0.006) & 0.633 (0.010) & 0.612 (0.031) & 0.813 (0.021) & 12           \\
TGN-m                        & 0.941 (0.004) & 0.857 (0.000) & 0.599 (0.027) & 0.643 (0.008) & 0.700 (0.044) & 0.493 (0.016) & 0.820 (0.070) & 7            \\
JODIE-h                      & 0.966 (0.004) & 0.965 (0.000) & 0.724 (0.006) & 0.888 (0.000) & 0.897 (0.014) & 0.682 (0.052) & 0.993 (0.001) & 2            \\
\bottomrule
\end{tabular}
}
}

\resizebox{\textwidth}{!}{
\subtable[Historical Sampling]{
\begin{tabular}{lcccccccc}
\toprule
Method                       & Wikipedia     & Reddit        & MOOC          & LastFM        & Enron         & Social Evo.   & UCI           & Rank         \\ \midrule
JODIE-s                      & 0.421 (0.001) & 0.411 (0.002) & 0.339 (0.001) & 0.432 (0.000) & 0.438 (0.001) & 0.418 (0.016) & 0.457 (0.017) & 12 \\
DyRep-s                      & 0.434 (0.002) & 0.465 (0.000) & 0.459 (0.022) & 0.464 (0.008) & 0.582 (0.022) & 0.621 (0.033) & 0.440 (0.009) & 7  \\
TGN-s                        & 0.690 (0.027) & 0.633 (0.000) & 0.552 (0.041) & 0.559 (0.029) & 0.601 (0.014) & 0.565 (0.045) & 0.603 (0.034) & 5  \\
JODIE-m                      & 0.421 (0.001) & 0.412 (0.002) & 0.337 (0.002) & 0.431 (0.000) & 0.438 (0.000) & 0.409 (0.011) & 0.436 (0.012) & 13 \\
DyRep-m                      & 0.427 (0.006) & 0.476 (0.000) & 0.502 (0.057) & 0.451 (0.001) & 0.589 (0.011) & 0.631 (0.023) & 0.435 (0.014) & 9  \\
TGN-m                        & 0.679 (0.005) & 0.699 (0.000) & 0.564 (0.042) & 0.626 (0.033) & 0.601 (0.015) & 0.512 (0.044) & 0.553 (0.106) & 4  \\
JODIE-h                      & 0.420 (0.000) & 0.417 (0.000) & 0.336 (0.003) & 0.430 (0.000) & 0.437 (0.000) & 0.450 (0.022) & 0.432 (0.023) & 14 \\
\bottomrule
\end{tabular}
}
}

\resizebox{\textwidth}{!}{
\subtable[Inductive Sampling]{
\begin{tabular}{lcccccccc}
\toprule
Method                       & Wikipedia     & Reddit        & MOOC          & LastFM        & Enron         & Social Evo.   & UCI           & Rank \\ \midrule
JODIE-s                      & 0.400 (0.000) & 0.441 (0.003) & 0.377 (0.024) & 0.458 (0.000) & 0.469 (0.001) & 0.423 (0.015) & 0.497 (0.012) & 11           \\
DyRep-s                      & 0.589 (0.014) & 0.522 (0.000) & 0.565 (0.037) & 0.523 (0.003) & 0.586 (0.017) & 0.617 (0.034) & 0.492 (0.009) & 5            \\
TGN-s                        & 0.667 (0.019) & 0.644 (0.000) & 0.532 (0.029) & 0.543 (0.029) & 0.631 (0.032) & 0.555 (0.035) & 0.637 (0.023) & 4            \\
JODIE-m                      & 0.401 (0.001) & 0.442 (0.003) & 0.363 (0.011) & 0.458 (0.000) & 0.469 (0.000) & 0.413 (0.009) & 0.482 (0.010) & 12           \\
DyRep-m                      & 0.602 (0.010) & 0.522 (0.000) & 0.544 (0.024) & 0.507 (0.008) & 0.582 (0.013) & 0.620 (0.023) & 0.486 (0.011) & 8            \\
TGN-m                        & 0.675 (0.013) & 0.709 (0.000) & 0.542 (0.009) & 0.558 (0.007) & 0.631 (0.032) & 0.527 (0.050) & 0.605 (0.080) & 3            \\
JODIE-h                      & 0.399 (0.000) & 0.448 (0.000) & 0.366 (0.015) & 0.457 (0.001) & 0.467 (0.002) & 0.457 (0.026) & 0.480 (0.017) & 14           \\
\bottomrule
\end{tabular}
}
}

\end{table}

\end{document}